%% file: main.tex
\pdfoutput=1
\documentclass[11pt,twoside]{article}
\usepackage{fancyhdr}
\usepackage[colorlinks,citecolor=blue,urlcolor=blue,linkcolor=blue,bookmarks=false]{hyperref}
\usepackage{amsfonts,epsfig,graphicx}
\usepackage{afterpage}
\usepackage{comment}
\usepackage{amsmath,amssymb,amsthm} 
\usepackage{fullpage}
\usepackage[T1]{fontenc} 
\usepackage{epsf} 
\usepackage{graphics} 
\usepackage{amsfonts,amsmath}
\usepackage[sort]{natbib} 
\usepackage{psfrag,xspace}
\usepackage{color,etoolbox}
\usepackage{subcaption}

\usepackage[utf8]{inputenc} 
\usepackage[T1]{fontenc}    
\usepackage{url}            
\usepackage{booktabs}       
\usepackage{amsfonts}       
\usepackage{nicefrac}       
\usepackage{microtype}      
\usepackage{amsthm}
\usepackage{amsmath}
\usepackage{amssymb}
\usepackage{xcolor}
\usepackage{mathtools}
\usepackage{comment}
\usepackage[noend]{algorithmic}
\usepackage[Algorithm]{algorithm}
\usepackage{float}
\usepackage{xr}
\usepackage{xspace}

\usepackage[suppress]{color-edits}
\addauthor[Amanda]{am}{blue}
\addauthor[Alex]{ac}{purple}
\addauthor[Edward]{ek}{green}

\theoremstyle{definition}
\newtheorem{theorem}{Theorem}[section]
\newtheorem{proposition}{Proposition}[section]
\newtheorem{corollary}{Corollary}[section]

\newtheorem{definition}{Definition}[section]
\newtheorem{condition}{Condition}[definition]

\newcommand{\E}{\mathbb{E}}
\newcommand{\En}{\hat{\mathbb{E}}_n}
\newcommand{\p}{\mathbb{P}}
\newcommand{\muo}{\mu}
\newcommand{\muoh}{\hat \mu}
\newcommand{\pih}{\hat \pi}
\newcommand{\plm}{\hat \nu_{\mathrm{PL}} (v)}
\newcommand{\plmg}{\hat \nu_{\mathrm{PL}}}

\newcommand{\bcm}{\hat \nu_{\mathrm{DR}} (v)}

\newcommand{\btrm}{\hat \nu_{\mathrm{TCR}} (v)}
\newcommand{\btro}{\omega(v)}
\newcommand{\ya}{Y^a}
\newcommand{\yo}{Y^0}
\newcommand{\indA}{\mathbb{I}\{A = a\}}
\newcommand\norm[1]{\left\lVert#1\right\rVert}
\newcommand{\vzcor}{\rho}

\newcommand{\kmu}{k_{\mu}}
\newcommand{\knu}{k_{\nu}}
\newcommand{\kpi}{k_{\pi}}
\newcommand{\kmuz}{k_{z}}
\newcommand{\kmuv}{k_{v}}
\newcommand{\kom}{k_{\omega}}

\newcommand{\dv}{d_\mathrm{V}}
\newcommand{\dz}{d_\mathrm{Z}}

\newcommand{\bc}{DR\xspace}
\newcommand{\pl}{PL\xspace}
\newcommand{\btr}{TCR\xspace}

\newcommand{\Wfirst}{\mathcal{W}^1}

\title{Counterfactual Predictions under Runtime Confounding}

\author{
 Amanda Coston \\
  Heinz College \& Machine Learning Dept.\\
  Carnegie Mellon University\\
  \texttt{acoston@cs.cmu.edu}
  \and
  Edward H. Kennedy \\
  Department of Statistics \\
  Carnegie Mellon University\\
 \texttt{edward@stat.cmu.edu} 
  \and
  Alexandra Chouldechova \\
  Heinz College \\
  Carnegie Mellon University\\
  \texttt{achould@cmu.edu}
}

\begin{document}

\maketitle

\begin{abstract}
  Algorithms are commonly used to predict outcomes under a particular decision or intervention, such as predicting likelihood of default if a loan is approved.
 Generally, to learn such \emph{counterfactual} prediction models from observational data on historical decisions and corresponding outcomes, one must measure all factors that jointly affect the outcome and the decision taken.
  Motivated by decision support applications, we study the counterfactual prediction task in the setting where all relevant factors are captured in the historical data, but it is infeasible, undesirable, or impermissible to use some such factors in the prediction model.
  We refer to this setting as \textbf{runtime confounding}.
  We propose a doubly-robust procedure for learning counterfactual prediction models in this setting. 
  Our theoretical analysis and experimental results suggest that our method often outperforms competing approaches.
  We also present a validation procedure for evaluating the performance of counterfactual prediction methods.

\end{abstract}

\section{Introduction}
\setcitestyle{numbers}
Algorithmic tools are increasingly prevalent in domains such as health care, education, lending, criminal justice, and child welfare \cite{caruana2015intelligible, smith2012predictive, khandani2010consumer, kehl2017algorithms, chouldechova2018case}.   In many cases, the tools are not intended to replace human decision-making, but rather to distill rich case information into a simpler form, such as a risk score, to inform human decision makers \cite{bezemer2019human, de2020case}.  The type of information that these tools need to convey is often \textit{counterfactual} in nature.  
Decision-makers need to know what is likely to happen if they choose to take a particular action.  For instance, an undergraduate program advisor determining which students to recommend for a personalized case management program might wish to know the likelihood that a given student will graduate if enrolled in the program.  
In child welfare, case workers and their supervisors may wish to know the likelihood of positive outcomes for a family under different possible types of supportive service offerings.

A common challenge to developing valid counterfactual prediction models is that all the data available for training and evaluation is observational: the data reflects historical decisions and outcomes under those decisions rather than randomized trials intended to assess outcomes under different policies.  If the data is confounded---that is, if there are factors not captured in the data that influenced both the outcome of interest and historical decisions---valid counterfactual prediction may not be possible.\accomment{I'm saying ``may not'' because I'm not ruling out the possibility of things like instruments existing that would still enable one to draw valid inferences.  What do you think?} \amcomment{I think this is reasonable. Another alternative would be "is not possible without further assumptions"}  In this paper we consider the setting where all relevant factors are captured in the data, and so historical decisions and outcomes are unconfounded, but where it is infeasible, undesirable, or impermissible to use some such factors in the prediction model.   We refer to this setting as \textbf{runtime confounding}.  

Runtime confounding naturally arises in a number of different settings.
First, relevant factors may not yet be available at the desired runtime.
For instance, in child welfare screening, call workers decide which allegations coming in to the child abuse hotline should be investigated based on the information in the call and historical administrative data \citep{chouldechova2018case}.
The call worker's decision-making process can be informed by a risk assessment if the call worker can access the risk score in real-time. Since existing case management software cannot run speech/NLP models in realtime, the call information (although recorded) is not available at runtime, thereby leading to runtime confounding.
Second, runtime confounding arises when historical decisions and outcomes have been affected by sensitive or protected attributes which for legal or ethical reasons are deemed ineligible as inputs to algorithmic predictions.
We may for instance be concerned that call workers implicitly relied on race in their decisions, but it would not be permissible to include race as a model input.
Third, runtime confounding may result from interpretability or simplicity requirements. 
For example, a university may require algorithmic tools used for case management to be interpretable. 
While information conveyed during student-advisor meetings is likely informative both of case management decisions and student outcomes, natural language processing models are not classically interpretable, and thus the university may wish instead to only use structured information like GPA in their tools.

\amcomment{
Runtime confounding naturally arises in a number of different settings.
First, relevant factors may not yet be available at the desired runtime.
Consider an in-person parole hearing, where the parole board makes a recommendation after reviewing documents and hearing spoken testimony.  The testimony may provide information that both influences the board's decision and reveals drivers of the offender's likelihood to succeed if released.  But the parole board would generally wish to see the predictions of a criminal risk and needs assessment tool prior to the hearing. Any such tool could not therefore use spoken testimony as a model input, thereby leading to runtime confounding.
Second, runtime confounding arises when historical decisions and outcomes have been affected by sensitive or protected attributes which for legal or ethical reasons are deemed ineligible as inputs to algorithmic predictions.
We may for instance be concerned that parole boards implicitly relied on race in their decisions, but it would not be permissible to include race as a model input. 
Third, runtime confounding may result from interpretability or simplicity requirements. 
For example, a university may require algorithmic tools used for case management to be interpretable. 
While information conveyed during student-advisor meetings is likely informative both of case management decisions and student outcomes, natural language processing models are not classically interpretable, and thus the university may wish instead to only use structured information like GPA in their tools.}

In practice, when it is undesirable or impermissible to use particular features as model inputs at runtime, it is common to discard the ineligible features from the training process.  This can induce considerable bias in the resulting prediction model when the discarded features are significant confounders.  
To our knowledge, the problem of learning valid counterfactual prediction models under runtime confounding has not been considered in the prior literature, leaving practitioners without the tools to properly incorporate runtime-ineligible confounding features into the training process.

\textbf{Contributions:} Drawing upon techniques used in low-dimensional treatment effect estimation \citep{van2003unified, zimmert2019nonparametric, chernozhukov2018generic}, we propose a procedure for the full pipeline of learning and evaluating prediction models under runtime confounding.
We
(1) formalize the problem of counterfactual prediction with runtime confounding [\textsection~\ref{sec:notation}];
(2) propose a solution based on doubly-robust techniques that has desirable theoretical properties [\textsection~\ref{sec:dr}];
(3) theoretically and empirically compare this solution to an alternative counterfactually valid approach as well as the standard practice, describing the conditions under which we expect each to perform well [\textsection~\ref{sec:methods} \&~\ref{sec:experiments}]; and (4) provide an evaluation procedure to assess performance of the methods in the real-world [\textsection~\ref{sec:eval}].  Proofs, code and results of additional experiments are presented in the Supplement. 

\subsection{Related work}

Our work builds upon a growing literature on counterfactual risk assessments for decision support that proposes methods for the unconfounded prediction setting \cite{schulam2017reliable, coston2020counterfactual}.
Following this literature, our goal is to predict outcomes under a proposed decision (interchageably referred to as `treatment' or `intervention') in order to inform human decision-makers about what is likely to happen under that treatment.

Our proposed prediction (Contribution 2) and evaluation methods (Contribution 4) draw upon the literature on double machine learning and doubly-robust estimation, which uses the efficient influence function to produce estimators with reduced bias
\citep{van2003unified, robins1994estimation, robins1995semiparametric, kennedy2016semiparametric, chernozhukov2018double, kennedy2020optimal}.
These techniques are commonly used for treatment effect estimation, and of particular note for our setting are methods for estimating treatment effects conditional on only a subset of confounders \citep{semenova2017estimation, chernozhukov2018generic,  zimmert2019nonparametric, fan2020estimation}. 
\citet{semenova2017estimation} propose a two-stage doubly-robust procedure that uses series estimators in the second stage to achieve asymptotic normality guarantees. \citet{zimmert2019nonparametric} propose a similar approach that uses local constant regression in the second stage, and \citet{fan2020estimation} propose using a local linear regression in the second stage. These approaches can obtain rate double-robustness under the notably strict condition that the product of nuisance errors to converge faster than $\sqrt{n}$ rates.  
In a related work, \citet{foster2019orthogonal} proposes an orthogonal estimator of treatment effects  which, under certain conditions, guarantees the excess risk is second-order but not doubly robust.\footnote{A second order but not doubly robust guarantee requires sufficiently fast rates on \emph{both} nuisance functions. By contrast, rate double robustness imposes a weaker assumption on the \emph{product} of nuisance function errors, allowing e.g., fast rates on the propensity function and slow rates on the outcome regression function.}
Our work is most similar to the approach taken in \citet{kennedy2020optimal}, which proposes a model-agnostic two-stage doubly robust estimation procedure for conditional average treatment effects that attains a model-free doubly robust guarantee on the prediction error.  
 Treatment effects can be identified under weaker assumptions than required to individual the potential outcomes, and prior work has proposed a procedure to find the minimal set of confounders for estimating conditional treatment effects \citep{makar2019distillation}.

Our prediction task is different from the common causal inference problem of treatment effect estimation, which targets a contrast of  outcomes under two different treatments \citep{wager2018estimation,shalit2017estimating}.
Treatment effects are useful for describing responsiveness to treatment.  While responsiveness is relevant to some types of decisions, it is insufficient, or even irrelevant, to consider for others.  
For instance, a doctor considering an invasive procedure  may make a different recommendation for two patients with the same responsiveness if one has a good probability of successful recovery without the procedure and the other does not.
In lending settings, the responsiveness to different loan terms is irrelevant; all that matters is that the likelihood of default be sufficiently small under feasible terms. In such settings, we are interested in \textit{predictions} conditional on only those features that are permissible or desirable to consider at runtime.
Our methods are specifically designed for minimizing prediction error, rather than providing inferential guarantees such as confidence intervals, as is common in the treatment effect estimation setting.

The practical challenge that we often need to make decisions based on only a subset of the confounders has been discussed in the policy learning literature \citep{zhang2012robust, athey2017efficient, kitagawa2018should}. For instance, it may be necessary to use only a subset of confounders to meet ethical requirements, model simplicity desiderata, or budget limitations  \citep{athey2017efficient}.
Doubly robust methods for learning treatment assignment policies have been proposed for such settings
\citep{zhang2012robust, athey2017efficient}. 


Our work is also related to the literature on marginal structure models (MSMs) \citep{robins2000amarginal,robins2000bmarginal}. An MSM is a model for a marginal mean of a counterfactual, possibly conditional on a subset of baseline covariates. The standard MSM approach is semiparametric, employing parametric assumptions for the marginal mean but leaving other components of the data-generating process unspecified \citep{van2003unified}. Nonparametric variants were studied in the unconditional case for continuous treatments by \citet{rubin2006extending}. In contrast our setting can be viewed as a nonparametric MSM for a binary treatment, conditional on a large subset of covariates. This is similar in spirit to partly-conditional treatment effect estimation \citep{van2014targeted}; however we do not target a contrast since our interest is in predictions rather than treatment effects. Our results are also less focused on model selection \citep{van2003bunified}, and more on error rates for particular estimators.
We draw on techniques for sample-splitting and cross-fitting, which have been used in the regression setting for model selection and tuning \citep{gyorfi2006distribution, van2003unified} and in treatment effect estimation
\citep{robins2008higher, zheng2010asymptotic, chernozhukov2018generic}.

Our method is relevant to settings where the outcome is selectively observed. This \emph{selective labels} problem \citep{lakkaraju2017selective, kleinberg2018human} is common in settings like lending where the repayment/default outcome is only observed for applicants whose loan is approved. Runtime confounding can arise in such settings if some factors that are used for decision-making are unavailable for prediction.

Recent work has considered methods to accommodate confounding due to sources other than missing confounders at runtime.
A line of work has considered how to use causal techniques to correct runtime dataset shift \cite{subbaswamy2018preventing, magliacane2018domain, subbaswamy2018counterfactual}.
In our case the runtime setting is different from the training setting not because of distributional shift but because we can no longer access all confounders.  These methods also differ from ours in that they are not seeking to predict outcomes under specific decisions.

There is also a line of work that considers confounding in the \emph{training} data
\cite{kallus2018confounding, madras2019fairness}.
While confounded training data is common in various applications, our work targets decision support settings where the factors used by decision-makers are recorded in the training data but are not available for prediction. 

Lastly, there are connections between runtime confounding and the literature on privileged learning and algorithmic fairness that use features during training time that are not available for prediction.
Learning using Privileged Information (LUPI)  has been proposed for settings in which the training data contains additional features that are not available at runtime \citep{vapnik2009new}. 
In algorithmic fairness, disparate learning processes (DLPs) use the sensitive attribute during training to produce models that achieve a target notion of parity without requiring access to the protected attribute at test time \cite{lipton2018does}.
LUPI and DLPs both make use of variables that are only available at train time, but if these variables affect the decisions under which outcomes are observed, predictions from LUPI and DLPs will be confounded because neither accounts for how these variables affect decisions. By contrast, our method uses confounding variables during training to produce valid counterfactual predictions.



\section{Problem setting} \label{sec:notation}
Our goal is to predict outcomes under a proposed treatment $A=a \in \{0,1\}$ based on runtime-available predictors  $V \in \mathcal{V} \subseteq \mathbb{R}^{\dv}$.\footnote{For exposition, we focus on making predictions for a single binary treatment $a$.   
To make predictions under multiple discrete treatments, our method can be repeated for each treatment using a one-vs-all setup.} 
Using the potential outcomes framework \citep{rubin2005causal, neyman1923applications},
our prediction target is ${\nu_a(v) := \E[\ya \mid V = v]}$  where $\ya \in \mathcal{Y} \subseteq \mathbb{R}$ is the potential outcome we would observe under treatment $A = a$.
We let $Z \in \mathcal{Z} \subseteq \mathbb{R}^{\dz}$ denote the runtime-hidden confounders, and we denote the propensity to receive treatment $a$ by ${\pi_a(v, z) := \p(A = a \mid V = v, Z = z)}$.
We also define the outcome regression by  ${\mu_a(v, z) := \E[\ya \mid V = v, Z = z]}$.
For brevity, we will generally omit the subscript, using notation $\nu$, $\pi$ and $\mu$ to denote the functions for a generic treatment $a$. 

\begin{definition} \label{def:problem_setting}
Formally, the task of counterfactual prediction under \textbf{runtime-only confounding} is to estimate ${\nu(v)}$ from iid training data $(V, Z, A, Y)$ under the following two conditions:
\begin{condition}[Training Ignorability]
\label{condition:ign}
Decisions are unconfounded given $V$ and $Z$: ${\ya \perp A \mid V, Z}$.
\end{condition}

\begin{condition}[Runtime Confounding] 
\label{condition:conf}
Decisions are confounded given only $V$: $\ya \not \perp A \mid V$; 
equivalently,  $A \not \perp Z \mid V$ and $\ya \not \perp Z \mid V$ 
\end{condition}
\end{definition}

To ensure that the target quantity is identifiable, we require two further assumptions, which are standard in causal inference and not specific to the runtime confounding setting.

\begin{condition}[Consistency]
\label{condition:consistency}
A case that receives treatment $a$ has outcome $Y = Y^a$.
\end{condition}

\begin{condition}[Positivity]
\label{condition:positivity}
$\p(\pi_a( V, Z)  \geq \epsilon > 0) = 1 \quad \forall a$ 
\end{condition}


\paragraph{Identifications.}
Under conditions 2.1.1-2.1.4, we can write the counterfactual regression functions $\mu$ and $\nu$ in terms of observable quantities. We can identify $\mu(v,z) = \E[Y \mid V = v, Z = z, A = a] $ and our target
$\nu(v) =\E [ \E[Y \mid V = v, Z = z, A = a] \mid V = v] = \E[ \muo(V, Z) \mid V = v]$. The iterated expectation in the identification of $\nu$ suggests a two-stage approach that we propose in \textsection~\ref{sec:pl} after reviewing current approaches.

\paragraph{Miscellaneous notation.} Throughout the paper we let $p(x)$ denote probability density functions; $\hat f$ denote an estimate of $f$; $L \lesssim R$ indicate that $L \leq C \cdot R$ for some universal constant $C$; $\mathbb{I}$ denote the indicator function; and define ${\norm{f}^2 := \int (f(x))^2 p(x) dx}$.  

\section{Prediction methods} \label{sec:methods}

\subsection{Standard practice: Treatment-conditional regression (\btr)} \label{sec:method_conf}
Standard counterfactual prediction methods train models on the cases that received treatment $a$  \citep{schulam2017reliable, coston2020counterfactual}, a procedure we will refer to as \textbf{treatment-conditional regression} (TCR).
This procedure estimates ${\btro = \E[Y \mid A = a, V =v]}$. 
This method works well given access to all the confounders at runtime; if ${A \perp Y^a \mid V}$, then ${\btro = \E[Y^a \mid V = v]} = \nu(v)$. 
However, under runtime confounding, ${\btro \neq \E[Y^a \mid V = v]}$, so this method does not target the right counterfactual quantity, and may produce misleading predictions.\footnote{Runtime imputation of $Z$ will not eliminate this bias since ${E[Y \mid A = a, V =v, f(v)] = \btro}$.}
For instance, consider a risk assessment setting that historically assigned
risk-mitigating treatment to cases that have higher risk under the null treatment ($A=0$).
Using \btr to predict outcomes under the null treatment will underestimate risk since ${\E[Y \mid V, A= 0] = \E[Y^0 \mid V, A= 0]  < \E[Y^0 \mid V]}$. 
We can characterize the bias of this approach by analyzing $b(v) := \btro -  \nu(v)$, a quantity we term the pointwise \emph{confounding bias.}
\begin{proposition} \label{prop:btr-bias}
Under runtime confounding, 
$\btro$ has pointwise confounding bias 
\begin{equation}
 b(v)  = \int_{\mathcal{Z}} \muo(v,z) \Big(p(z \mid V =  v, A = a) - p(z \mid V = v) \Big) dz \quad \neq \quad 0
\end{equation}
\end{proposition}
By Condition~\ref{condition:conf}, this confounding bias will be non-zero. 
Nonetheless we might expect the \btr method to perform well if $b(v)$ is small enough.  
We can formalize this intuition by decomposing the error of a \btr predictive model $\hat{\nu}_{\mathrm{TCR}}$ into estimation error and confounding bias:

\begin{proposition}
\label{thm:tcr}
The pointwise regression error of the TCR method can be bounded as follows:
\begin{align*}
    \E[(\nu(v) - \btrm)^2] &\lesssim \E[(\btro - \btrm)^2] + b(v)^2
\end{align*}
\end{proposition}
The first term gives the estimation error and the second term bounds the bias in targeting the wrong counterfactual quantity. 

\subsection{A simple proposal: Plug-in (\pl) approach}
\label{sec:pl}

\begin{algorithm}[t]
\caption{The plug-in (\pl) approach}
\label{alg:pl}
\begin{algorithmic}
  \scriptsize
  \STATE \emph{Stage 1:} Learn $\muoh(v,z)$ by regressing $Y \sim V, Z \mid A = a$  
  \STATE \emph{Stage 2:} Learn $\plm$ by regressing $\muoh(V,Z) \sim V$
\end{algorithmic}
\end{algorithm}

We can avoid the confounding bias of TCR through a simple two-stage procedure we call the \textbf{plug-in} approach that targets the proper counterfactual quantity.
This approach, described in Algorithm~\ref{alg:pl}, first estimates $\muo$ and then uses $\mu$ to construct a pseudo-outcome which is regressed on $V$ to yield prediction $\hat{\nu}_{\mathrm{PL}}$. 
Cross-fitting techniques (Alg.~\ref{alg:pl_cf}) can be applied to prevent issues that may arise due to potential overfitting when learning both $\muoh$ and $\plmg$ on the same training data.
Sample-splitting (or cross-fitting) also enables us to get the following upper bound on the error of the \pl{} approach.

\begin{algorithm}[t]
\caption{The plug-in (\pl) approach with cross-fitting}
\label{alg:pl_cf}
\begin{algorithmic}
  \scriptsize
  \STATE Randomly divide training data into two partitions $\mathcal{W}^1$ and  $\mathcal{W}^2$.
  \FOR{ $(p,q) \in \{(1,2), (2,1)\}$}
  \STATE \emph{Stage 1:} On partition $\mathcal{W}^p$, learn $\muoh^p(v,z)$ by regressing $Y \sim V, Z \mid A = a$  
  \STATE \emph{Stage 2:} On partition $\mathcal{W}^{q}$, learn $\hat \nu^q_{\mathrm{PL}}(v)$ by regressing $\muoh^p(V,Z) \sim V$
  \ENDFOR
  \STATE \textbf{\pl{} prediction:} $\plm = \frac{1}{2} \sum_{i =1}^2 \hat{\nu}^i_{\mathrm{PL}}(v)$ 
\end{algorithmic}
\end{algorithm}

\begin{proposition}\label{thm:pl}
Under sample-splitting for stages 1 and 2 and stability conditions on the 2nd stage estimators (appendix), the \pl{} method has pointwise regression error bounded by
\begin{align*}
    \E \Big[\big(\plm - \nu(v)\big)^2 \Big] \lesssim & \  \E \Big[\big(\tilde \nu(v) - \nu(v)\big)^2\Big] +  \E \Big[ \big(\muoh(V, Z) - \muo(V, Z)\big)^2 \mid V = v \Big]
\end{align*}
where the oracle-quantity $\tilde \nu(v)$
describes the function we would get in the second-stage if we had oracle access to $\ya$. 
\end{proposition}
This simple approach can consistently estimate our target $\nu(v)$.
However, it solves a harder problem (estimation of $\mu(v,z)$)  than what our lower-dimensional target $\nu$ requires. 
Notably the bound depends \emph{linearly} on the MSE of $\hat \mu$. We next propose an approach that avoids such strong dependence.


\subsection{Our main proposal: Doubly-robust (\bc) approach} \label{sec:dr}
Our main proposed method is what we call the \textbf{doubly-robust} (\bc) approach, which improves upon the \pl{} procedure by using a bias-corrected pseudo-outcome in the second stage (Alg.~\ref{alg:dr}).
The \bc{} approach estimates both $\mu$ and $\pi$, which enables the method to perform well in situations in which $\pi$ is easier to estimate than $\mu$.
 We propose a cross-fitting (Alg.~\ref{alg:dr_cf}) variant that satisfies the sample-splitting requirements of Theorem~\ref{thm:bc}.

 \begin{algorithm}[t] 
\caption{The proposed doubly-robust (\bc) approach}
\label{alg:dr_cf}
\begin{algorithmic}
  \scriptsize
  \STATE \emph{Stage 1:} Learn $\muoh(v,z)$ by regressing $Y \sim V, Z \mid A = a$.
  \\ \quad \qquad \ \  Learn $\pih(v,z)$ by regressing $\mathbb{I}\{A=a\} \sim V, Z$ 
  \STATE \emph{Stage 2:}  Learn $\bcm$ by regressing $\Big( \frac{\mathbb{I}\{A=a\}}{\pih(V,Z)}(Y - \muoh(V,Z)) + \muoh(V,Z) \Big) \sim V$ 
\end{algorithmic}
\end{algorithm}

 \begin{algorithm}[t] 
\caption{The proposed doubly-robust (\bc) approach with cross fitting}
\label{alg:dr}
\begin{algorithmic}
  \scriptsize
  \STATE  Randomly divide training data into  three partitions $\mathcal{W}^1$, $\mathcal{W}^2$, $\mathcal{W}^3$.
  \FOR{$(p, q, r) \in \{(1,2,3), (3,1,2), (2,3,1)\}$}
  \STATE \emph{Stage 1:} On $\mathcal{W}^p$, learn $\muoh^p(v,z)$ by regressing $Y \sim V, Z \mid A = a$.
  \\ \quad \qquad \ \  On $\mathcal{W}^{q}$, learn $\pih^{q}(v,z)$ by regressing $\mathbb{I}\{A=a\} \sim V, Z$ 
  \STATE \emph{Stage 2:}  On $\mathcal{W}^{r}$, learn $\hat{\nu}_{\mathrm{DR}}^r$ by regressing $\Big( \frac{\mathbb{I}\{A=a\}}{\pih^{q}(V,Z)}(Y - \muoh^p(V,Z)) + \muoh^p(V,Z) \Big) \sim V$ 
  \ENDFOR
  \STATE \textbf{\bc{} prediction:} $\bcm = \frac{1}{3}  \sum_{i =1}^3 \hat{\nu}_{\mathrm{DR}}^i(v)$
\end{algorithmic}
\end{algorithm}

\begin{theorem} \label{thm:bc}
Under sample-splitting to learn $\muoh$, $\pih$, and $\hat{\nu}_{\mathrm{DR}}$ and stability conditions on the 2nd stage estimators (appendix), the \bc{} method has pointwise error bounded by:
\begin{align*} \label{eqn:bc-rate}
  \E \Big[\big(\bcm - \nu(v)\big)^2 \Big] \lesssim \ & \E \Big[\big(\tilde \nu(v) - \nu(v)\big)^2\Big] \\
  &+ \E \Big[ (\pih(V, Z) - \pi(V, Z))^2 \mid V = v\Big] \E \Big[ (\muoh(V, Z) - \muo(V, Z))^2 \mid V = v\Big]
\end{align*}

This implies a similar bound on the integrated MSE (given in appendix).

\end{theorem}
The \bc error is bounded by the error of an oracle with access to $\ya$ and a \emph{product} of nuisance function errors.\footnote{The term \emph{nuisance} refers to functions $\mu$ and $\pi$.}
 This product can be substantially smaller than the error of $\hat \mu$ in the \pl bound.
  When this product is less than the oracle error, the \bc{} approach is oracle-efficient, in the sense that it achieves (up to a constant factor) the same error rate as an oracle.
This model-free result provides bounds that hold for \emph{any} regression method.
It is nonetheless instructive to consider the form of these bounds in a couple common contexts. The next result is specialized to the sparse high-dimensional setting, and subsequently we consider the smooth non-parametric setting. 
\begin{corollary} \label{cor:lasso}
Assume stability conditions on the 2nd stage regression estimator (appendix) and that a $k$-sparse model can be estimated with squared error $k^2 \sqrt{\frac{\log d}{n}}$ (e.g. \cite{chatterjee2013assumptionless}).\footnote{We use the sparsity parameter $k$ to indicate $k$ covariates have non-zero coefficients in the model.}
With $\kom$-sparse $\omega$, the pointwise error for the \btr method is 
\begin{align*}
  \E \Big[\big(\btrm - \nu(v)\big)^2 \Big] \lesssim &\ \   \kom^2 \sqrt{\frac{\log \dv}{n}} + b(v)^2
\end{align*}
With $\kmu$-sparse $\mu$ and $\knu$-sparse $\nu$,
 the pointwise error for the \pl{} method is 
\begin{align*}
  \E \Big[\big(\plm - \nu(v)\big)^2 \Big] \lesssim & \ \  \knu^2 \sqrt{\frac{\log \dv}{n}} + \kmu^2\sqrt{\frac{\log d}{n}} 
\end{align*}

Additionally with $\kpi$-sparse $\pi$, the pointwise error for the \bc{} method is 
\begin{align*}
  \E \Big[\big(\bcm - \nu(v)\big)^2 \Big] \lesssim &\ \   \knu^2 \sqrt{\frac{\log \dv}{n}} + \kmu^2\kpi^2 \frac{\log d}{n} 
\end{align*}

The \bc approach is therefore oracle efficient when $\Big(\frac{\kmu \kpi}{\knu}\Big)^2 \lesssim \Big(\frac{\sqrt{n \log\dv}}{\log d} \Big)$.
\end{corollary}
Based on the upper bound, we cannot claim efficiency for the \pl approach because 
$k_\mu > k_\nu$ and $d > \dv$.
For exposition, consider the simple case where $k_{\nu} \approx k_{\mu} \approx k_{\pi}$.
Corollary~\ref{cor:lasso} indicates that when $\dv \approx d$, the \bc{} and \pl{} methods will perform similarly. 
When $\dv \ll d$, we expect the \bc{} to outperform the \pl{} method because the second term of the PL bound dominates the error whereas the first term of the DR bound dominates in high-dimensional settings. 
When $\dv \ll d$ and the amount of confounding is small, we expect the \btr to perform well.

\begin{corollary} \label{cor:smooth}
Assume stability conditions on the 2nd stage regression estimator (appendix) and that a $\beta$-smooth function of a $p$-dimensional vector can be estimated with squared error $n^{\frac{-2\beta}{2\beta + p}}$.
With $\beta_{\omega}$-smooth $\omega$, the pointwise error for the \btr method is 
\begin{align*}
  \E \Big[\big(\btrm - \nu(v)\big)^2 \Big] \lesssim &\ \   n^{-2\beta_{\omega}/(2\beta_{\omega} + \dv)} + b(v)^2
\end{align*}
With $\beta_{\mu}$-smooth $\mu$ and $\beta_{\nu}$-smooth $\nu$,
 the pointwise error for the \pl{} method is 
\begin{align*}
  \E \Big[\big(\plm - \nu(v)\big)^2 \Big] \lesssim & \ \  n^{-2\beta_{\nu}/(2\beta_{\nu} + \dv)} + n^{-2\beta_{\mu}/(2\beta_{\mu} + d)}  
\end{align*}

Additionally with $\beta_{\pi}$-smooth $\pi$, the pointwise error for the \bc{} method is 
\begin{align*}
  \E \Big[\big(\bcm - \nu(v)\big)^2 \Big] \lesssim &\ \   n^{-2\beta_{\nu}/(2\beta_{\nu} + \dv)} + n^{\frac{-2\beta_{\mu}}{2\beta_{\mu} + d} + \frac{-2\beta_{\pi}}{2\beta_{\pi} + d}} 
\end{align*}

The \bc approach is therefore oracle efficient when $\frac{\beta_{\nu}}{\beta_{\nu} + \dv/2} \leq \frac{\beta_{\mu}}{\beta_{\mu} + d/2} + \frac{\beta_{\pi}}{\beta_{\pi} + d/2}$ which simplifies to $s \geq \frac{d/2}{1 + \frac{\dv}{\beta_{\nu}}}$ when $\beta_{\pi} = \beta_{\mu} = s$.
\end{corollary}

As in the sparse setting above, we cannot claim oracle efficiency for \pl approach based on this upper bound because
$\beta_\mu \leq \beta_\nu$ and $d > \dv$.
For exposition, consider an example where $\beta_{\nu} \approx \beta_{\mu} \approx \beta_{\pi}$.
Corollary~\ref{cor:smooth} indicates that when $\dv \approx d$, the \bc{} and \pl{} methods will perform similarly. 
When $\dv \ll d$, we expect the \bc{} to outperform the \pl{} method because the second term of the PL bound dominates the error whereas the first term of the DR bound dominates. 
When $\dv \ll d$ and the amount of confounding is small, we expect the \btr to perform well.

This theoretical analysis helps us understand when we expect the prediction methods to perform well.
However, in practice, these upper bounds may not be tight and the degree of confounding is typically unknown. 
To compare the prediction methods in practice, we require a method for counterfactual model evaluation.

\section{Evaluation method} \label{sec:eval}
We describe an approach for evaluating the prediction methods using observed data.
In our problem setting 
(\textsection~\ref{def:problem_setting}),
the prediction error of a model $\hat \nu$ is identified as $\E[(\ya - \hat \nu(V))^2] = \E[\mathbb{E}[(Y-\hat{\nu}(V))^2 \mid V, Z, A= a]]$.
We propose a doubly-robust procedure to estimate the prediction error that follows the approach in \cite{coston2020counterfactual}, which focused on classification metrics and therefore did not consider MSE.
Defining the error regression $\eta(v, z) \coloneqq \E[(\ya-\hat{\nu}(V))^2|V = v, Z = z]$, which is identified as $\mathbb{E}[(Y-\hat{\nu}(V))^2 \mid V = v, Z = z, A= a]$, 
the \textbf{doubly-robust estimate of the MSE of $\nu$} is
\begin{equation*}
    \frac{1}{n}\sum_{i=1}^{n}\Bigg[\frac{\mathbb{I}\{A_i = a\}}{\pih(V_i,Z_i)} \Big(\big(Y_i-\hat{\nu}(V_i)\big)^2 - \hat{\eta}(V_i,Z_i)\Big) + \hat{\eta}(V_i,Z_i)\Bigg]
\end{equation*}
The doubly-robust estimation of MSE is $\sqrt{n}$-consistent under sample-splitting and $n^{1/4}$ convergence in the nuisance function error terms, enabling us to get estimates with confidence intervals. Algorithm~\ref{alg:eval} describes this procedure.\footnote{The appendix describes a cross-fitting approach to jointly learn and evaluate the three prediction methods.}
This evaluation method can also be used to select the regression estimators for the first and second stages.

\begin{algorithm}[t]
\caption{Cross-fitting approach to evaluation of counterfactual prediction methods}
\label{alg:eval}
\begin{algorithmic}
  \scriptsize
  \STATE \textbf{Input:} Test samples $\{(V_j, Z_j, A_j, Y_j)\}_{j=1}^{2n}$ and prediction models $\{\hat{\nu}_1, ... \hat{\nu}_h\}$
  \STATE  Randomly divide test data into two partitions $\mathcal{W}^0 = \{(V^0_j, Z^0_j, A^0_j, Y^0_j)\}_{j=1}^n$ and $\mathcal{W}^1 = \{(V^1_j, Z^1_j, A^1_j, Y^1_j)\}_{j=1}^n$.
  \FOR{$(p, q) \in \{(0,1), (1,0)\}$}
  \STATE On $\mathcal{W}^p$, learn $\pih^p(v,z)$ by regressing $\mathbb{I}\{A=a\} \sim V, Z$. 
  \FOR{$m \in \{1,...., h\}$}
  \STATE  On $\mathcal{W}^p$, learn $\hat \eta^p_{m}(v,z)$ by regressing $(Y-\hat{\nu}_{m}(V))^2 \sim V, Z \mid A= a$
  \STATE On $\mathcal{W}^q$, for $j \in \{1,..., n\}$  compute ${\phi^q_{m, j} =  \frac{\mathbb{I}\{A^q_j = a\}}{\pih^{p}(V^q_j,Z^q_j)} ((Y^q_j-\hat{\nu}_{m}(V^q_j))^2 - \hat{\eta}^p_{m}(V^q_j,Z^q_j)) + \hat{\eta}^p_{m}(V^q_j,Z^q_j)}$
  \ENDFOR
  \ENDFOR
    \STATE  \textbf{Output error estimate confidence intervals:} for $m \in \{ 1,..., h \}$: \\ $\mathrm{MSE}_m = \Big(\frac{1}{2n} \displaystyle \sum_{i =0}^1 \displaystyle \sum_{j=1}^n  \phi^i_{m, j} \Big) \pm 1.96 \sqrt{\frac{1}{2n} \mathrm{var}(\phi_{m})}$ 
\end{algorithmic}
\end{algorithm}


\section{Experiments} \label{sec:experiments}

We evaluate our methods against ground truth by performing experiments on simulated data, where we can vary the amount of confounding in order to assess the effect on predictive performance.
While our theoretical results for \pl{} and \bc{} are obtained under sample splitting, in practice there may be a reluctance to perform sample splitting in training predictive models due to the potential loss in efficiency.  
In this section we present results where we use the full training data to learn the 1st-stage nuisance functions and 2nd-stage regressions for \bc{} and \pl{} and we use the full training data for the one-stage \btr.\footnote{We report error metrics on a random heldout test set.}
This allows us to examine performance in a setting outside what our theory covers.

\begin{figure}
\centering
      \includegraphics[width=0.2\linewidth]{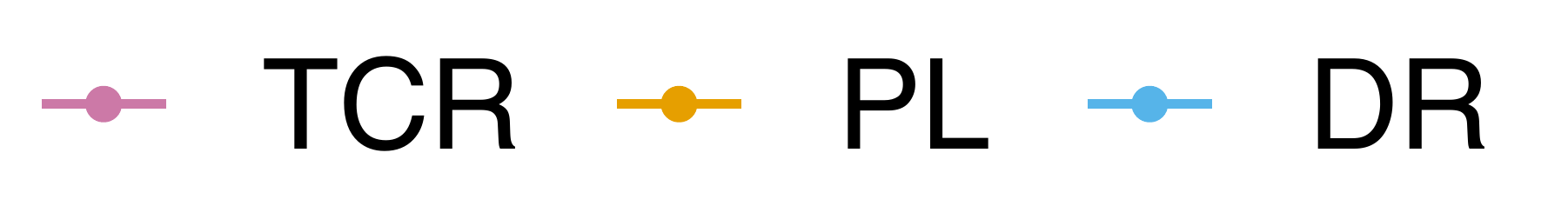}\\
      \includegraphics[width=0.03\textwidth, trim=300 0 220 0]{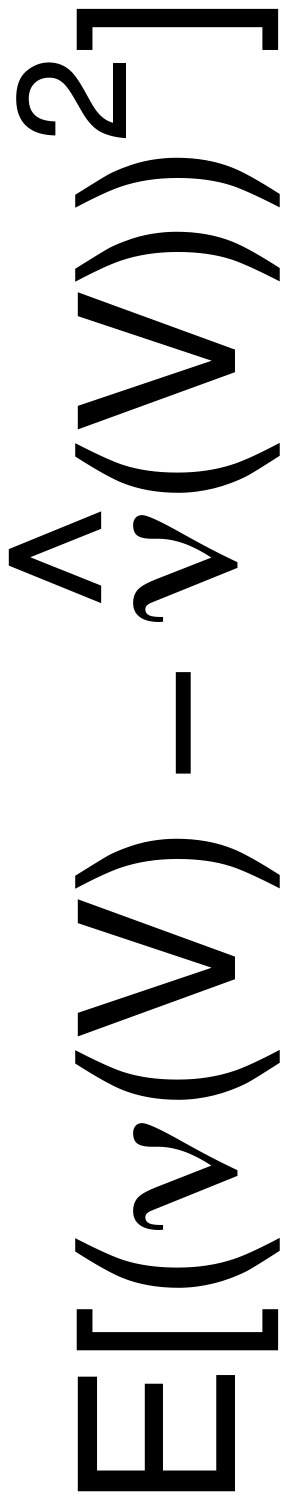}
    \begin{subfigure}[t]{0.31\textwidth}
        \centering
        \includegraphics[width=\linewidth, trim=23 20 14 15]{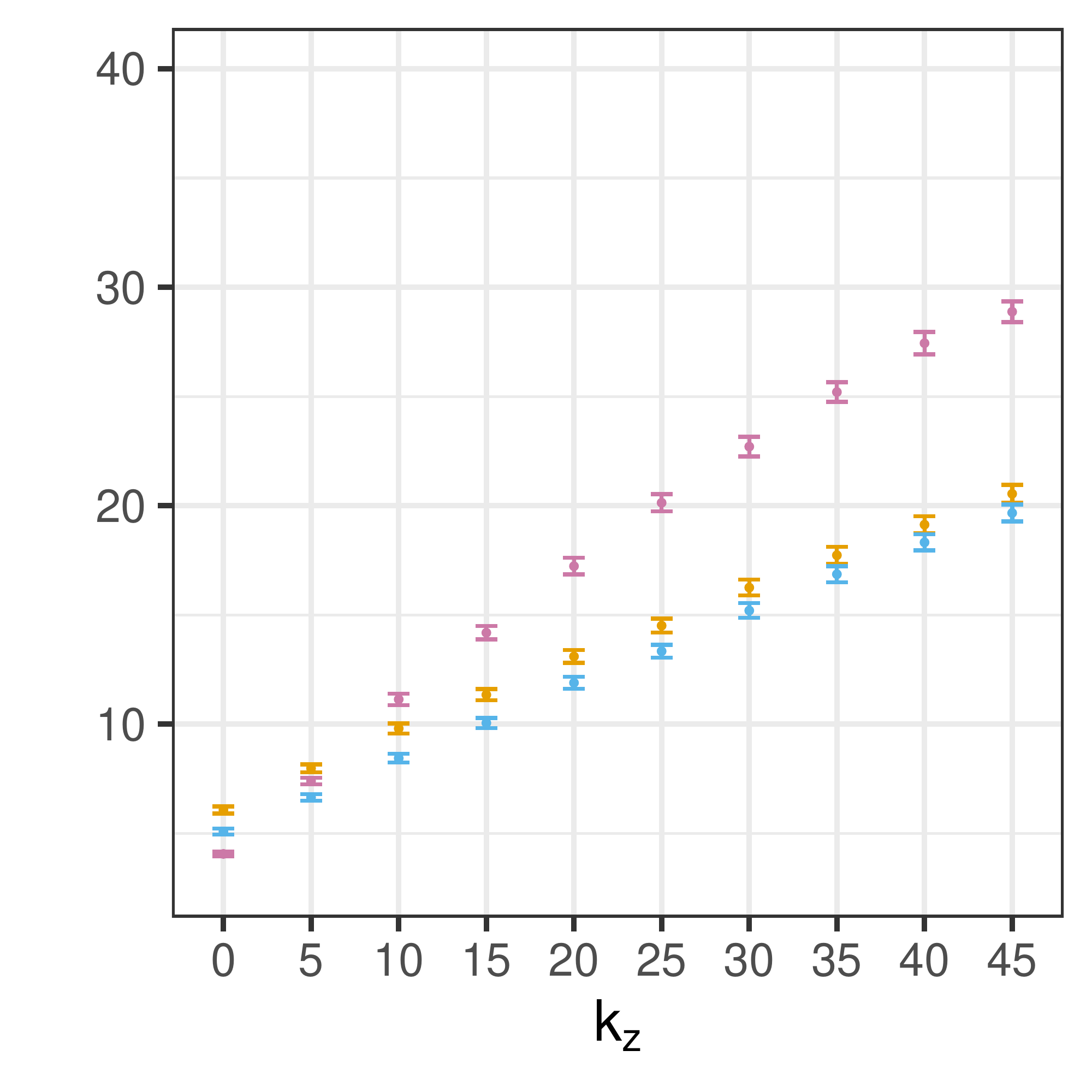}
        \caption{LASSO}
        \end{subfigure}%
    ~
    \begin{subfigure}[t]{0.31\textwidth}
        \centering
         \includegraphics[width=\linewidth, trim=23 20 14 15]{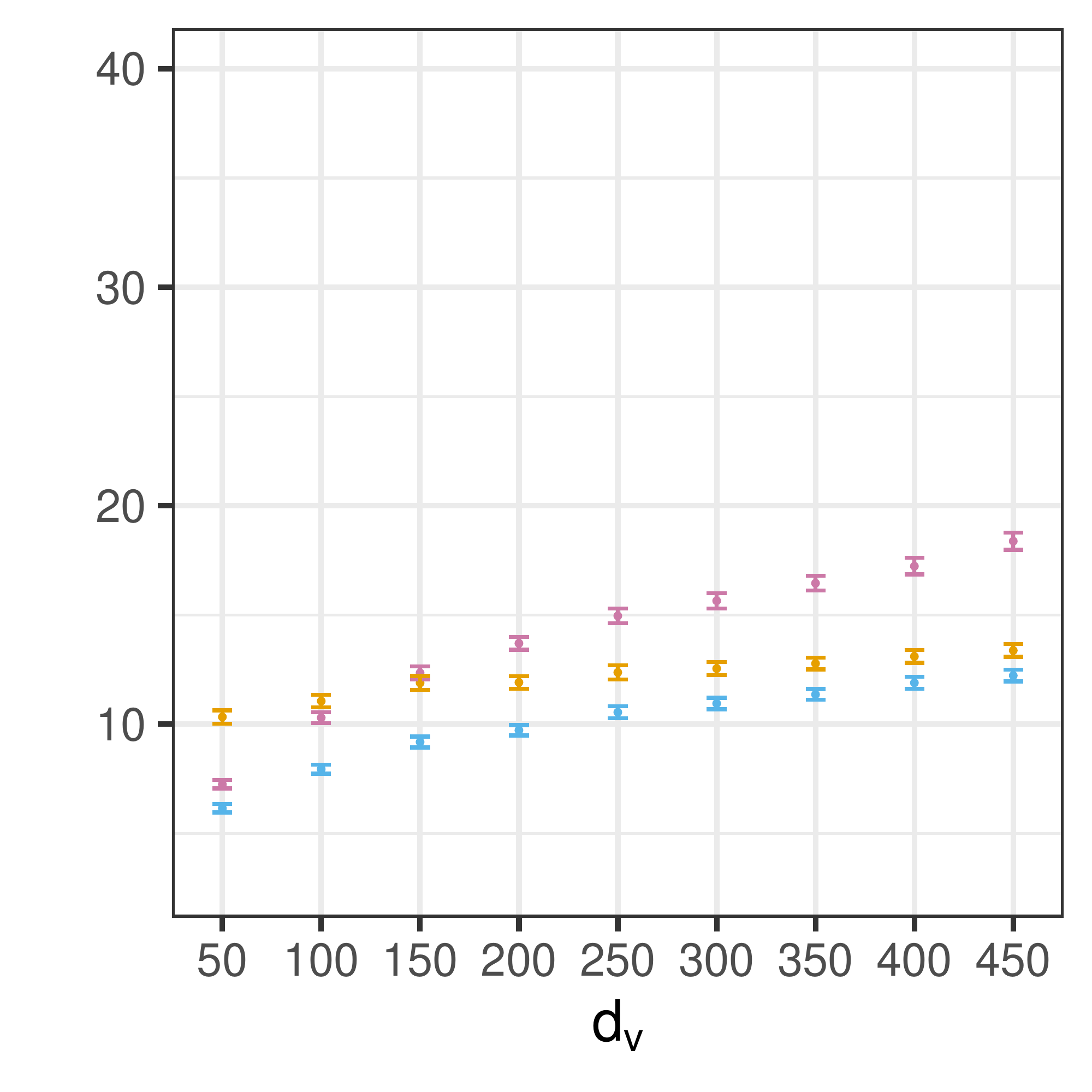}
        \caption{LASSO}
    \end{subfigure}%
    ~
    \begin{subfigure}[t]{0.31\textwidth}
        \centering
        \includegraphics[width=\linewidth, trim=23 20 14 15]{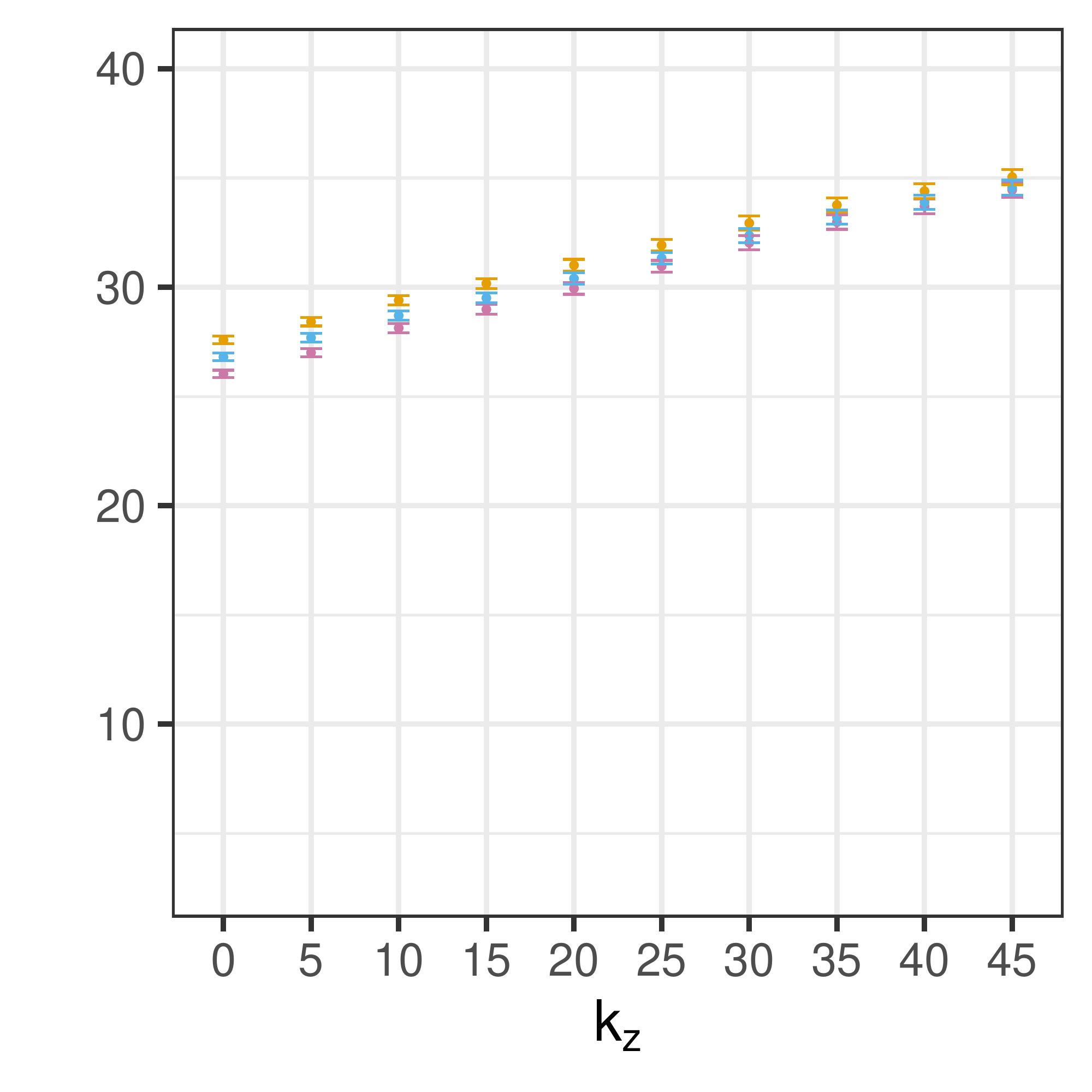}
         \caption{Random forests}
    \end{subfigure}
    \caption{\textbf{(a)} MSE as we vary $\kmuz$ using cross-validated LASSO to learn $\pih$, $\muoh$, $\hat{\nu}_{\mathrm{TCR}}$, $\hat{\nu}_{\mathrm{PL}}$, $\hat{\nu}_{\mathrm{DR}}$ for $\vzcor = 0$,  $\dv = 400$ and $\kmuv = 25$. At low levels of confounding ($\kmuz$), the \btr method does well but performance degrades with $\kmuz$. For any non-zero confounding, our \bc{} method performs best. \\
    \textbf{(b)} MSE against $\dv$ using cross-validated LASSO and $\vzcor = 0$,  $\kmuv = 25$ and $\kmuz = 20$. The \bc{} method performs the best across the range of $\dv$.
    When $\dv$ is small, the \btr method also does well since its estimation error is small.
    The \pl{} method has higher error since it suffers from the full $d$-dimensional estimating error in the first stage. 
    \textbf{(c)} MSE as we vary $\kmuz$ using random forests and $\vzcor = 0$,  $\dv = 400$ and $\kmuv = 25$. Compared to LASSO in (a), there is a relatively small increase in error as we increase $\kmuz$, suggesting that estimation error dominates the confounding error. 
    The \btr method performs best at lower levels of confounding and on par with the \bc{} method for larger values of $\kmuz$. \\
     Error bars denote $95\%$ confidence intervals.}
    \label{fig:uncor}
\end{figure}

We first analyze how the methods perform in a sparse linear model. 
This simple setup enables us to explore how properties like correlation between $V$ and $Z$ impact performance. 
We simulate data as
\begin{align*}
    V_i & \sim \mathcal{N}(0,1) \hspace{50pt} ;\ 1 \leq i \leq \dv\\
    Z_i & \sim \mathcal{N}(\rho V_i, 1-\rho^2) \hspace{19pt} ;\ 1 \leq i \leq \dz\\
    \muo(V, Z)  &= \frac{\kmuv}{\kmuv+\rho \kmuz} \Big(\sum_{i =1}^{\kmuv} V_i + \sum_{i =1}^{\kmuz} Z_i \Big) \hspace{28pt}  
    \ \    \ya = \muo(V,Z) + \epsilon \hspace{3pt} ;\ \epsilon \sim \mathcal{N}\Bigg(0, \frac{1}{2n}\norm{\muo(V,Z)}_2^2 \Bigg) \\
    \nu(V) &=   \frac{\kmuv}{\kmuv+\rho \kmuz}\Big( \sum_{i =1}^{\kmuv} V_i +  \rho  \sum_{i =1}^{\kmuz} V_i \Big) \\
      \pi(V, Z) &= 1 - \sigma\Bigg(\frac{1}{\sqrt{\kmuv + \kmuz}} \Big( \sum_{i=1}^{\kmuv} V_i + \sum_{i=1}^{\kmuz} Z_i \Big) \Bigg) \hspace{20pt}  \ \ 
      A \sim \mathrm{Bernoulli}(\pi(V,Z)) 
\end{align*}
where $\sigma(x) = \frac{1}{1 + e^{-x}}$. 
We normalize $\pi(v,z)$ by $\frac{1}{\sqrt{ \kmuv + \kmuz}}$ to satisfy Condition~\ref{condition:positivity} and use the coefficient $\kmuv/(\kmuv+\rho \kmuz)$ to facilitate a fair comparison as we vary $\rho$.
For all experiments, we report test MSE for 300 simulations where each simulation generates $ n = 2000$ data points split randomly and evenly into train and test sets.\footnote{Source code is available at \href{https://github.com/mandycoston/confound_sim}{https://github.com/mandycoston/confound\_sim}}
In the first set of experiments, for fixed $d = \dv + \dz = 500$, we vary $\dv$ (and correspondingly $\dz$). 
 We also vary $\kmuz$, which governs the runtime confounding.
Larger values of $\kmuz$ correspond to more confounding variables.
The theoretical analysis (\textsection~\ref{sec:methods}) suggests that when  confounding ($\kmuz$) is small, then the \btr and \bc{} methods will perform well. 
More confounding (larger $\kmuz$) should increase error for all methods, and we expect this increase to be significantly larger for the \btr method that has confounding bias.
We expect the \btr and \bc{} methods to perform better at smaller values of $\dv$; by contrast, we expect the \pl{} performance to vary less with $\dv$ since the \pl{} method suffers from the full $d$-dimensionality in the first stage regardless of $\dv$.
For large values of $\dv$, we expect the \pl{} method to perform similarly to the \bc{} method. 
Fig.~\ref{fig:uncor} plots the MSE in estimating $\nu$ for $\vzcor = 0$ and $\kmuv = 25$ using LASSO and random forests.
The LASSO plots in Fig.~\ref{fig:uncor}a and~\ref{fig:uncor}b show the expected trends.
Random forests have much higher error than the LASSO (compare Fig.~\ref{fig:uncor}a to~\ref{fig:uncor}c) and we only see a small increase in error as we increase confounding (Fig.~\ref{fig:uncor}c) because the random forest estimation error dominates the confounding error.
In this setting, the \btr method may outperform the other methods, and in fact the \btr performs best at low levels of confounding.

\begin{figure}
\centering
      \includegraphics[width=0.2\linewidth]{img/legend.pdf}\\
      \includegraphics[width=0.03\textwidth, trim=300 0 220 0]{img/axis.pdf}
    \begin{subfigure}[t]{0.31\textwidth}
        \centering
        \includegraphics[width=\linewidth, trim=23 20 14 15]{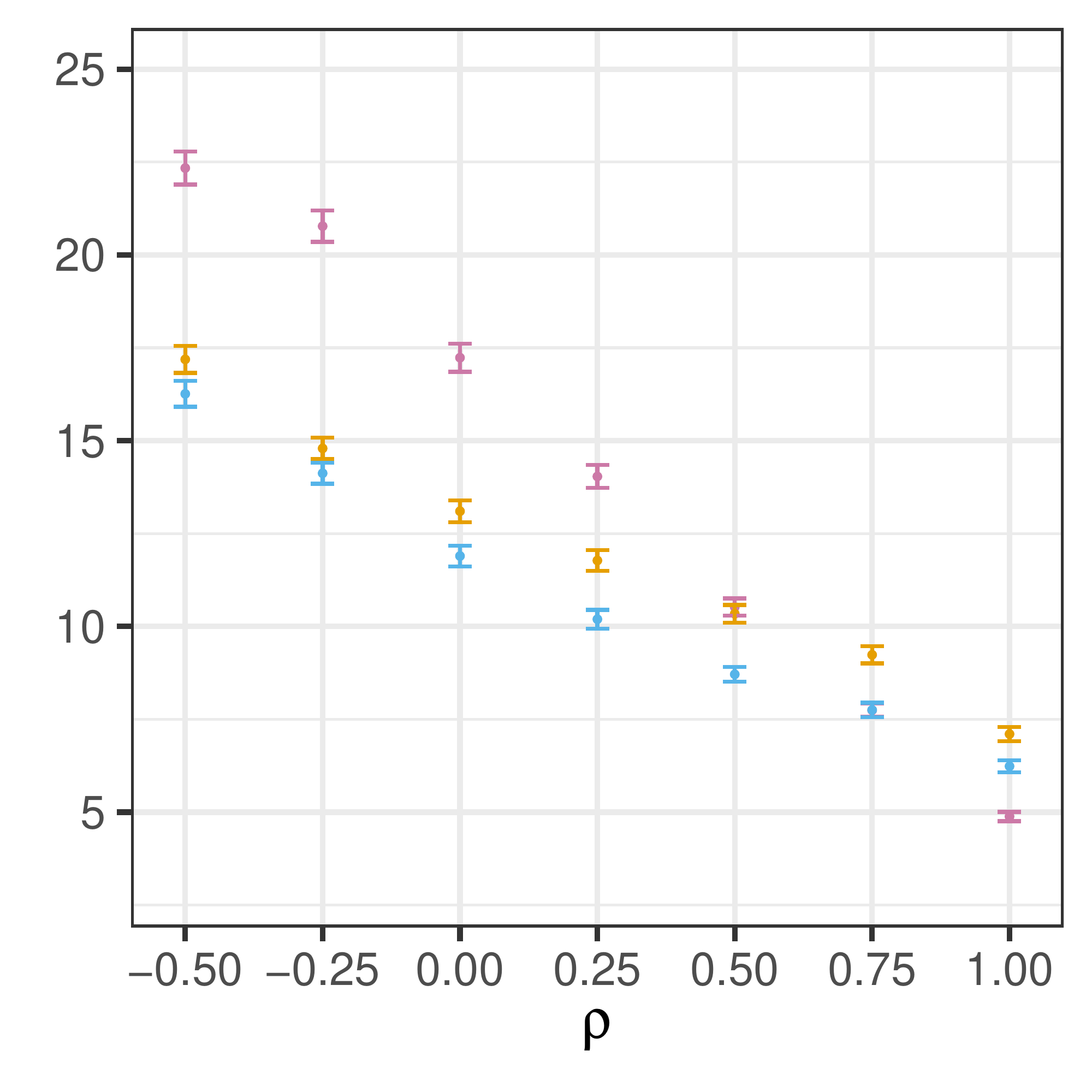}
        \caption{LASSO}
        \end{subfigure}%
    ~
    \begin{subfigure}[t]{0.31\textwidth}
        \centering
         \includegraphics[width=\linewidth, trim=23 20 14 15]{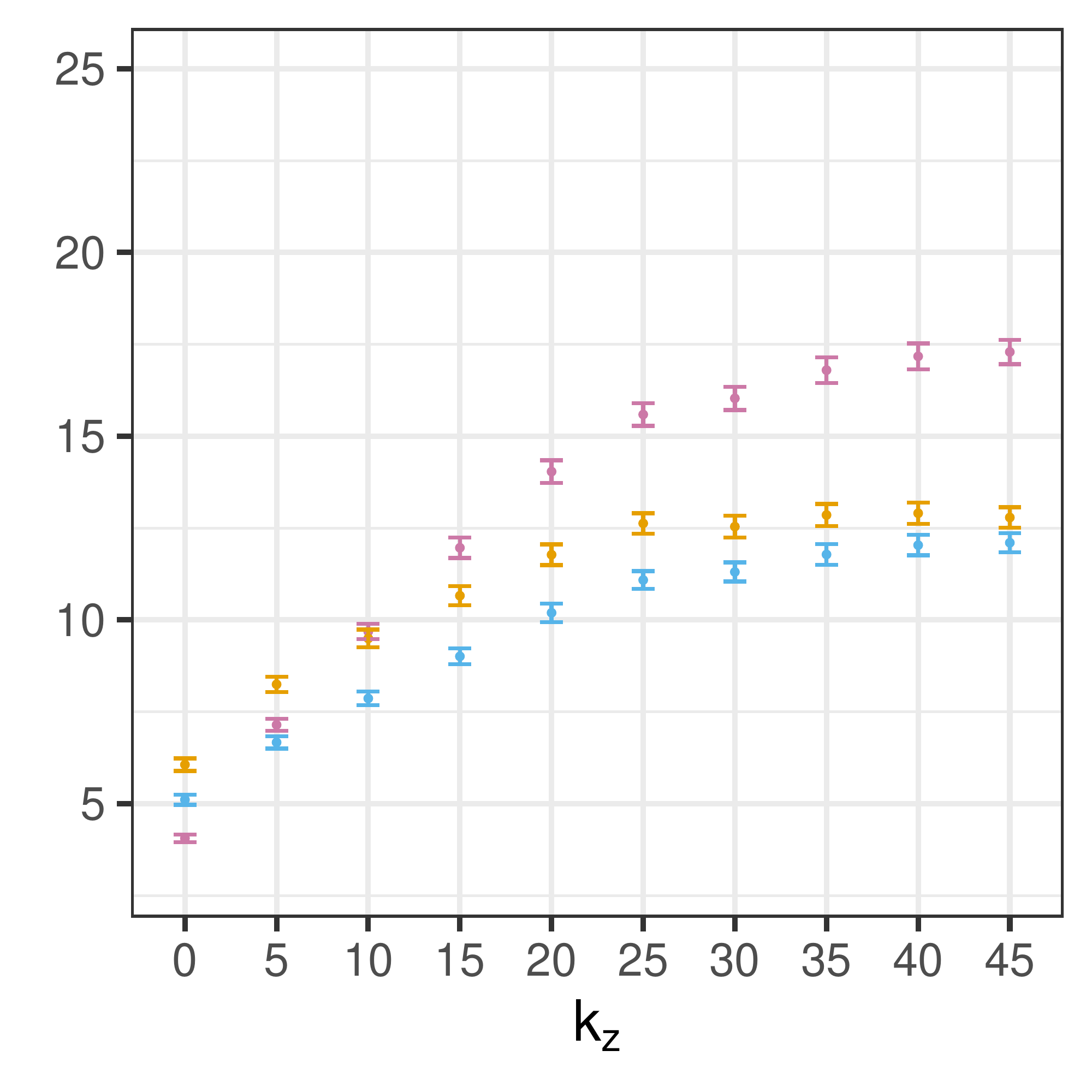}
        \caption{LASSO}
    \end{subfigure}%
    ~
    \begin{subfigure}[t]{0.31\textwidth}
        \centering
        \includegraphics[width=\linewidth, trim=23 20 14 15]{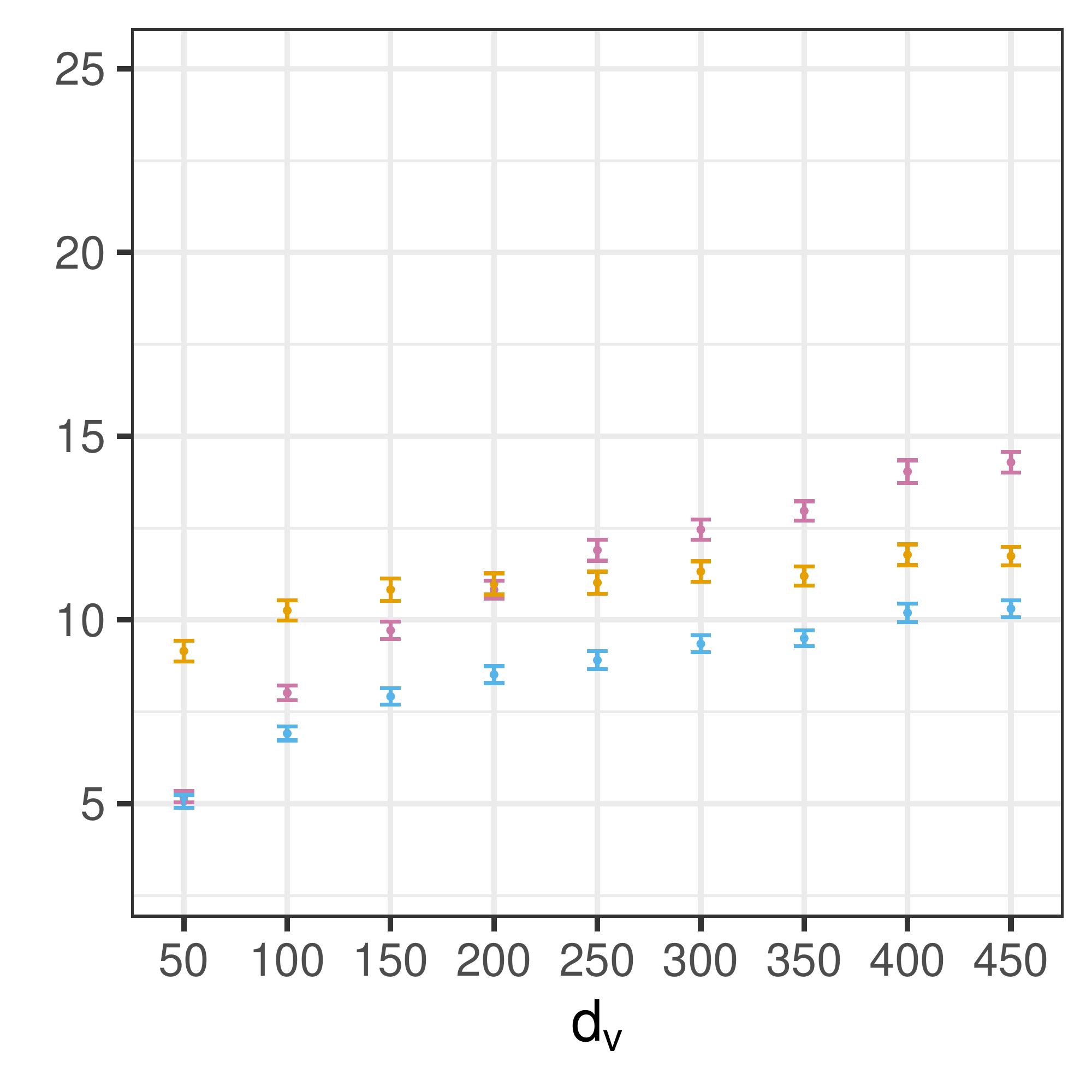}
         \caption{LASSO}
    \end{subfigure}
    \caption{\textbf{(a)} MSE against correlation $\vzcor_{V_i, Z_i}$ for $\kmuz = 20$, $\kmuv = 25$, and $\dv = 400$.
Error decreases with $\vzcor$ for all methods.
Our \bc{} method achieves the lowest error under confounding  ($\vzcor < 1$).
    \textbf{ (b)} MSE as we increase $\kmuz$ for $\vzcor = 0.25$, $\kmuv = 25$, and $\dv = 400$. Compare to  Figure~\ref{fig:uncor}a; the weak positive correlation reduces MSE, particularly for  $\kmuv < i \leq \kmuz$ when $V_i$ is only a correlate for the confounder $Z_i$ but not a confounder itself. 
    \textbf{(c)} MSE against $\dv$ for $\vzcor = 0.25$, $\kmuz = 20$, and $\kmuv = 25$. The \bc{} method is among the best-performing for all $\dv$. 
As with the uncorrelated setting (\ref{fig:uncor}b), the \bc{} and \btr methods are better able to take advantage of low $\dv$ than the \pl{} method. \\
 Error bars denote $95\%$ confidence intervals.
}
    \label{fig:cor}
\end{figure}

We next consider the case were $V$ and $Z$ are correlated.
If V and Z are perfectly correlated, there is no confounding.
For our data where higher values of $V$ and $Z$ both decrease $\pi$ and increase $\mu$, 
a positive correlation should reduce confounding, and
a negative correlation may exacerbate confounding by increasing the probability that $Z$ is small given $A = a$ and $V$ is large and therefore increasing the gap $\E[\ya \mid V = v] - \E[\ya \mid V = v, A = a]$.
Fig.~\ref{fig:cor} gives MSE for correlated V and Z. 
As expected, error overall decreases with $\vzcor$ (Fig.~\ref{fig:cor}a).
Relative to the uncorrelated setting (Fig.~\ref{fig:uncor}), the weak positive correlation reduces MSE for all methods, particularly for large $\kmuz$ and $\dv$.
The \bc{} method achieves the lowest error for settings with confounding, performing on par with the \btr when $\dv = 50$.

\paragraph{Experiments with Second-Stage Misspecification}
Next, we explore a more complex data generating process through the lens of model interpretability.
Interpretability requirements allow for a complex training process as long as the final model outputs interpretable predictions \citep{tan2018distill, zeng2017interpretable, rudin2019stop}.  
Since the \pl{} and \bc{} first stage regressions are only a part of the training process, we can use any flexible model to learn the first stage functions as accurately as possible without impacting interpretability.
Constraining the second-stage learning class to interpretable models (e.g. linear classifiers) may cause misspecification since the interpretable class may not contain the true model. 
We simulate such a setting by modifying the setup (for $\vzcor = 0$): 
\begin{align*}
    V_i & \sim \mathcal{N}(0,1) \mathrm{\ \ for \ } \ 1 \leq i \leq \frac{\dv}{2} \hspace{15pt} 
    \\ V_i &\coloneqq V_j^2 \mathrm{\ \ for \ \ }  \frac{\dv}{2} < i \leq \dv, \hspace{7pt}  j = i - \frac{\dv}{2}\\
    \muo(V, Z) & = \sum_{i =1}^{\kmuv/2} \Big( V_i + (2(i \bmod  2) -1) V_i^2 \Big) + \sum_{i=1}^{\kmuz} Z_i \hspace{8pt} \\ 
    \nu(V) &=   \sum_{i =1}^{\kmuv/2} \Big( V_i + (2(i \bmod 2) -1) V_i^2 \Big)
\end{align*}
 We restrict our second stage models and the \btr model to predictors $V_i$ for $1 \leq i \leq \frac{\dv}{2}$ to simulate a real-world setting where we are constrained to linear classifiers using only $V$ at runtime.
 We allow the first stage models access to the full $V$ and $Z$ since the first stage is not constrained by variables or model class.
We use cross-validated LASSO models for both stages and compare this setup to the setting where the model is correctly specified. 
The \bc{} method achieves the lowest error for both settings (Table~\ref{fig:mse_misspec}), although the error is significantly higher for all methods under misspecification.

\begin{table}[ht]
\centering
\begin{tabular}{c}
  Mean-Squared Error $\E[\big(\nu(V) - \hat{\nu}(V)\big)^2]$ \\
\end{tabular}
\begin{tabular}{rlr}
  \hline
 Method & Correct specification & 2nd-stage misspecification  \\ 
  \hline
 TCR & 16.64 (16.28, 17.00) & 35.52 (35.18, 35.85) \\ 
  PL & 12.32 (12.03, 12.61) & 32.09 (31.82, 32.36) \\ 
  DR (ours) & \textbf{11.10 (10.84, 11.37)} & \textbf{31.33 (31.06, 31.59)} \\ 
   \hline
\end{tabular}
\caption{MSE $\E[\big(\nu(V) - \hat{\nu}(V)\big)^2]$ under correct specification vs misspecification in the 2nd stage for $d= 500$, ${\dv = 400}$, ${\kmuv = 24}$, ${\kmuz = 20}$ and $n=3000$ (with 95\% confidence intervals). Our DR method has the lowest error in both settings. Errors are larger for all methods under misspecification.} 
\label{fig:mse_misspec}
\end{table}

\subsection{Experiments on real-world child welfare data}
In the US, each year over 4 million calls are made to child welfare screening hotlines with allegations of child neglect or abuse \citep{usdhhs}. 
Call workers must decide which allegations coming in to the child abuse hotline should be investigated.
In agencies that have adopted risk assessment tools, the worker relies on (immediate risk) information communicated during the call and
an algorithmic risk score that summarizes (longer term) risk based on historical administrative data \citep{chouldechova2018case}.
The call is recorded but is not used as a predictor for three reasons: (1) the inadequacy of existing case management software to run speech/NLP models on calls in realtime; (2) model interpretability requirements; and (3) the need to maintain distinction between immediate risk (as may be conveyed during the call) and longer-term risk the model seeks to estimate.
Since it is not possible to use call information as a predictor, we encounter runtime confounding.
Additionally, we would like to account for the disproportionate involvement of families of color in the child welfare system \cite{dettlaff2011disentangling}, but due to its sensitivity, we do not want to use race in the prediction model. 

The task is to predict which cases are likely to be offered services under the decision $a = $ ``screened in for investigation'' using historical administrative data as predictors ($V$) and accounting for confounders race and allegations in the call ($Z$). Our dataset consists of over 30,000 calls to the hotline in Allegheny County, PA. We use random forests in the first stage for flexibility and LASSO in the second stage for interpretability.
Table~\ref{table:child_welfare} presents the MSE using our evaluation method (\textsection~\ref{sec:eval}).\footnote{We report error metrics on a random heldout test set.} The \pl and \bc methods achieve a statistically significant lower MSE than the \btr approach, suggesting these approaches could help workers better identify at-risk children than standard practice.

\begin{table}[ht]
\centering
\begin{tabular}{rlrrr}
  \hline
 & MSE  \\ 
  \hline
 \btr & 0.290 (0.287, 0.293) \\ 
 \pl  & \textbf{0.249 (0.246, 0.251)} \\ 
 \bc (ours) & \textbf{0.248 (0.245, 0.250)} \\ 
   \hline
\end{tabular}
\caption{MSE estimated via our evaluation procedure (\textsection~\ref{sec:eval}) for child welfare screening task. The \pl and \bc approaches achieve lower MSE than the \btr approach. $95\%$ confidence intervals given.} 
\label{table:child_welfare}
\end{table}


\section{Conclusion}
We propose a generic procedure for learning counterfactual predictions under runtime confounding that can be used with any parametric or nonparametric learning algorithm. 
Our theoretical and empirical analysis suggests this procedure will often outperform other methods, particularly when the level of runtime confounding is significant.

\section*{Acknowledgements}
This research was made possible through support from the Tata Consultancy Services (TCS) Presidential Fellowship, the K\&L Gates Presidential Fellowship, and the Block Center for Technology and Society at Carnegie Mellon University.
This material is based upon work supported by the National Science Foundation Grants No. DMS1810979, IIS1939606, and the National Science Foundation  Graduate
Research Fellowship Program under Grant No. DGE1745016. Any opinions,
findings, and conclusions or recommendations expressed in this material are those of the
author(s) and do not necessarily reflect the views of the National Science Foundation.
We are grateful to Allegheny County Department of Human Services for sharing their data.
Thanks to our reviewers for providing useful feedback about the project and to Siddharth Ancha for helpful discussions.

\clearpage

\bibliographystyle{plainnat}
\bibliography{ref}

\clearpage

\appendix
\input{supp}

\end{document}

%% file: supp.tex
\section{Details on Proposed Learning Procedure}

We describe a joint approach to learning and evaluating the \btr, \pl, and \bc prediction methods in Algorithm~\ref{alg:joint}. 
This approach efficiently makes use of the need for both prediction and evaluation methods to estimate the propensity score $\pi$.

\begin{algorithm} 
\caption{Cross-fitting procedure to learn and evaluate the \btr, \pl, and \bc prediction methods}
\label{alg:joint}
\begin{algorithmic}
  \scriptsize
  \STATE \textbf{Input:} Data samples $\{(V_j, Z_j, A_j, Y_j)\}_{j=1}^{4n}$
  \STATE  Randomly divide training data into four partitions $\mathcal{W}^1$, $\mathcal{W}^2$, $\mathcal{W}^3$, $\mathcal{W}^4$ where $\mathcal{W}^1 = \{(V^1_j, Z^1_j, A^1_j, Y^1_j)\}_{j=1}^n$ (and similarly for $\mathcal{W}^2$, $\mathcal{W}^3$, $\mathcal{W}^4$).
  \FOR{$(p,q,r,s) \in \{(1,2,3,4), (4,1,2,3), (3,4,1,2), (2,3,4,1)\}$}
  \STATE \emph{Stage 1:} On $\mathcal{W}^p$, learn $\muoh^p(v,z)$ by regressing $Y \sim V, Z \mid A = a$.
  \\ \quad \qquad \ \  On $\mathcal{W}^q$, learn $\pih^q(v,z)$ by regressing $\mathbb{I}\{A=a\} \sim V, Z$ 
  \STATE \emph{Stage 2:}  On $\mathcal{W}^r$, learn $\hat{\nu}^r_{\mathrm{DR}}$ by regressing $\Big( \frac{\mathbb{I}\{A=a\}}{\pih^q(V,Z)}(Y - \muoh^p(V,Z)) + \muoh^p(V,Z) \Big) \sim V$ 
  \\ \quad \qquad \ \ On $\mathcal{W}^r$ and $\mathcal{W}^q$, learn $\hat{\nu}^r_{\mathrm{PL}}$ by regressing $\muoh^p(V,Z) \sim V$
  \\ \quad \qquad \ \ On $\mathcal{W}^r$, $\mathcal{W}^q$, and $\mathcal{W}^p$, learn $\hat{\nu}^r_{\mathrm{TCR}}$ by regressing $Y \sim V \mid A = a$ 
    \STATE \emph{Evaluate} for $m$ in \{ \btr, \pl, \bc \}: 
    \\ \quad \qquad \ \ On $\mathcal{W}^q$, learn $\hat \eta^q_{m}(v,z)$ by regressing $(Y-\hat{\nu}^r_{m}(V))^2 \sim V, Z \mid A= a$ 
     \\ \quad \qquad \ \ On $\mathcal{W}^s$, for $j = 1,...n$  compute ${\phi^s_{m,j} =  \frac{\mathbb{I}\{A_j = a\}}{\pih^q(V_j,Z_j)} ((Y_j-\hat{\nu}^r_{m}(V_j))^2 - \hat{\eta}^q_{m}(V_j,Z_j)) + \hat{\eta}^q_{m}(V_j,Z_j)}$
  \ENDFOR
  \STATE \textbf{Output prediction models:} $\bcm = \frac{1}{4} \displaystyle \sum_{j =1}^4 \hat{\nu}_{\mathrm{DR}, j}(v))$; 
  \quad $\plm = \frac{1}{4}\displaystyle \sum_{j =1}^4  \hat{\nu}_{\mathrm{PL}, j}(v)$ ; 
  \quad  $\btrm = \frac{1}{4} 
 \displaystyle \sum_{j =1}^4 \hat{\nu}_{\mathrm{TCR}, j}(v)$
    \STATE  \textbf{Output error estimate confidence intervals:} for $m$ in \{ \btr, \pl, \bc \}: \\ $\mathrm{MSE}_m = \Big(\frac{1}{4n} \displaystyle \sum_{i =1}^4 \displaystyle \sum_{j=1}^n  \phi^i_{m,j} \Big) \pm 1.96 \sqrt{\frac{1}{4n} \mathrm{var}(\phi_{m})}$ 
\end{algorithmic}
\end{algorithm}

\section{Proofs and derivations}
 In this section we provided detailed proofs and derivations for all results in the main paper. 
 
\subsection{Derivation of Identifications of $\mu$ and $\nu$} \label{sec:ident}
We first show the steps to identify $\mu(v,z)$:
\begin{align*}
 \mu(v,z) &= \E[\ya \mid V = v, Z = z] \\
    \E[\ya \mid V = v, Z = z] &=   \E[\ya \mid V = v, Z = z, A = a]  \\
     &=  \E[Y \mid V = v, Z = z, A = a] \\
\end{align*}
The first line applies the definition of $\mu$..
The second  line follows from training ignorability (Condition~\ref{condition:ign}).
The third line follows from consistency (Condition~\ref{condition:consistency}). 

Next we show the identification of $\nu(v)$:
\begin{align*}
\nu(v) &= \E[\ya \mid V = v] \\
    \E[\ya \mid V = v] &= \E [ \E[\ya \mid V = v, Z = z] \mid V = v] \\
     &= \E [ \E[\ya \mid V = v, Z = z, A = a] \mid V = v] \\
     &=  \E [ \E[Y \mid V = v, Z = z, A = a] \mid V = v] \\
\end{align*}
The first line applies the definition of $\nu$ from Section~\ref{sec:notation}.
The second line follows from iterated expectation.
The third line follows from training ignorability (Condition~\ref{condition:ign}).
The fourth line follows  from consistency (Condition~\ref{condition:consistency}).

Note that we can concisely rewrite the last line as $\E[\mu(V,Z) \mid V = v]$ since we have identified $\mu$.

 \subsection{Proof that \btr method underestimates risk under mild assumptions on a risk assessment setting} \label{sec:proof_btr_under}
\begin{proof}
In Section~\ref{sec:method_conf} we posited that the \btr method will often underestimate risk in a risk assessment setting. 
We demonstrate this for the setting with a binary outcome $Y \in \{0,1\}$, but the logic extends to settings with a discrete or continuous outcome.
We assume larger values of $Y$ are adverse i.e. $Y = 0$ is desired and $Y = 1$ is adverse.
The decision under which we'd like to estimate outcomes is the baseline decision $A = 0$.
We start by recalling runtime confounding condition (\ref{condition:conf}): $\p(A= 0 \mid V, \yo =1) \neq \p(A= 0 \mid V, \yo =0)$. 
Here we further refine this by assuming we are in the common setting where treatment $A =1$ is more likely to be assigned to people who are higher risk.
Then $\p(A= 1 \mid V, \yo =1) > \p(A= 1 \mid V, \yo =0)$.
Equivalently $\p(A= 0 \mid V, \yo =1) < \p(A= 0 \mid V, \yo =0)$.
By the law of total probability,
$$\p(A = 0 \mid V) = \p(A = 0 \mid V, \yo = 1) \p(\yo = 1 \mid V) + \p(A = 0 \mid V, \yo = 0) \p(\yo = 0 \mid V)$$
Assuming $\p(\yo = 1 \mid V) > 0$, this implies 
\begin{equation} \label{eqn:btr_proof1}
    \p(A= 0 \mid V, \yo =0) > \p(A = 0 \mid V)
\end{equation}
By Bayes' rule,
$$\p(A= 0 \mid V, \yo =0)  = \p(\yo= 0 \mid V, A =0)\frac{ \p(A= 0 \mid V)}{\p(\yo = 0 \mid V)}  $$
Using this in the LHS of Equation~\ref{eqn:btr_proof1} and dividing both sides of Equation~\ref{eqn:btr_proof1} by $\p(A = 0 \mid V)$, we get 
$$ \frac{\p(\yo= 0 \mid V, A =0)}{\p(\yo = 0 \mid V)} > 1 $$  
Equivalently
 ${ \E[\yo \mid V, A= 0]  < \E[\yo \mid V]}$.
\end{proof}

\subsection{Derivation of Proposition~\ref{prop:btr-bias} (confounding bias of the \btr method)}
We recall \textbf{Proposition~\ref{prop:btr-bias}}: \\
Under runtime confounding, a model that perfectly predicts 
$\btro$ has pointwise confounding bias $b(v) = \omega(v) - \nu(v) =$
\begin{equation}
 \int_{\mathcal{Z}} \muo(v,z) \Big(p(z \mid V =  v, A = a) - p(z \mid V = v) \Big) dz \quad \neq \quad 0
\end{equation}

\begin{proof} 
By iterated expectation and the definition of expectation we have that 
\begin{align*}
    \omega(v) &= \int_{\mathcal{Z}}\E[Y \mid V = v , Z = z, A = a] \ p(z \mid V = v, A =a) dz  \\
    & = \int_{\mathcal{Z}}\mu(v,z) p(z \mid V = v, A =a) dz 
\end{align*}

In the identification derivation above we saw that $\nu(v) = \E[\mu(V,Z) \mid V = v]$.
Using the definition of expectation, we can rewrite $\nu(v)$ as 
\begin{align*}
&= \int_{\mathcal{Z}} \mu(v,z) p(z \mid V = v) dz \\
\end{align*}

Therefore the pointwise bias is
\begin{equation}
\omega(v) -  \nu(v) =   \int_{\mathcal{Z}} \mu(v,z) \Big(p(z \mid V = v, A = a) - p(z \mid V = v) \Big) dz 
\end{equation}

We can prove that this pointwise bias is non-zero by contradiction. Assuming the pointwise bias is zero, we have $\omega(v) = \nu(v) \implies \ya \perp A \mid V = v$ which contradicts the runtime confounding condition~\ref{condition:conf}. 
\end{proof}
We emphasize that the confounding bias does not depend on the treatment effect. 
This approach is problematic whenever treatment assignment depends on $Y^a$ to an extent that is not measured by $V$, even for settings with no treatment effect (such as selective labels setting \cite{lakkaraju2017selective, kleinberg2018human}).

\subsection{Proof of Proposition~\ref{thm:tcr} (error of the \btr method)} \label{sec:proof_thm_tcr}
We can decompose the pointwise error of the \btr method into the estimation error and the bias of the \btr target.
\begin{proof}
\begin{align*}
    \E[(\nu(v) - \btrm)^2] &=\E[\Big((\nu(v) - 
    \btro) + (\btro - \btrm )\Big)^2] \\ 
    &\leq 2\Bigg( \E[(\nu(v) - 
    \btro)^2] + \E[(\btro - \btrm )^2] \Bigg) \\
    &\lesssim (\nu(v) - 
    \btro)^2 + \E[(\btro - \btrm )^2] \\
    & = b(v)^2 + \E[(\btro - \btrm )^2]
\end{align*}
Where the second line is due to the fact that $(a+b)^2 \leq 2(a^2 + b^2)$.
In the third line, we drop the expectation on the first term since there is no randomness in two fixed functions of $v$. 
\end{proof}

\subsection{Proofs of Proposition~\ref{thm:pl} and Theorem~\ref{thm:bc} (error of the \pl and \bc methods)} \label{sec:proof_thms}
We begin with additional notation needed for the proofs of the error bounds.
For brevity let $W = (V, Z, A, Y)$ indicate a training observation. 
The theoretical guarantees for our methods rely on a two-stage training procedure that assumes independent training samples.
We denote the first-stage training dataset as $\mathcal{W}^1 := \{W^1_{1}, W^1_{2} ,W^1_{3}, ... W^1_{n} \}$ and the second-stage training dataset as $\mathcal{W}^2 := \{W^2_{1}, W^2_{2} ,W^2_{3}, ... W^2_{n} \}$.
Let $\En[Y \mid V = v]$ denote an estimator of the regression function $\E[Y \mid V = v]$.
Let $L \asymp R$ denote $L \lesssim \ R $  and $R \lesssim \ L$.

\begin{definition}(Stability conditions) \label{condition:stab}
The results assume the following two stability conditions on the second-stage regression estimators:
\begin{condition}
$\En[Y \mid V = v] + c = \En[Y + c \mid V = v]$ for any constant $c$
\end{condition}
\begin{condition}
For two random variables $R$ and $Q$,
if $\E[R \mid V = v] = \E[Q \mid V = v]$, then 
$$\E \Bigg[ \Big( \En[R \mid V = v] - \E[R \mid V = v] \Big)^2  \Bigg] \asymp \E \Bigg[ \Big( \En[Q \mid V = v] - \E[Q \mid V = v] \Big)^2  \Bigg]$$ 
\end{condition}
\end{definition}

The second condition is satisfied for instance by local estimation techniques. While global methods (such as linear regression) may not satisfy this property, a weaker stability condition (see \citet{kennedy2020optimal}) can be used  to achieve a bound on the integrated mean squared error.

\subsubsection{Proof of Proposition~\ref{thm:pl} (error of the \pl method)}
The theoretical results for our two-stage procedures rely on the theory for pseudo-outcome regression in \citet{kennedy2020optimal} which bounds the error for a two-stage regression on the full set of confounding variables.
However, our setting is different since our second-stage regression is on a subset of confounding variables. 
Therefore, Theorem 1 of \citet{kennedy2020optimal} does not immediately give the error bound for our setting, but we can use similar techniques in order to get the bound for our V-conditional second-stage estimators.
\begin{proof}
As our first step, we define an error function.
The error function of the \pl approach is  $\hat{r}_{\mathrm{PL}}(v)$
\begin{align*}
     &= \E[\muoh(V, Z) \mid V = v, \Wfirst] - \nu(v) \\
    &= \E[\muoh(V, Z) \mid V = v, \Wfirst] - \E[ \muo(V, Z) \mid V = v] \\
    &= \E[\muoh(V, Z) - \muo(V, Z) \mid V = v, \Wfirst] \\
\end{align*}
The first line is our definition of the error function (following \cite{kennedy2020optimal}).
The second line uses iterated expectation, and the third lines uses the fact that $\Wfirst$ is a random sample of the training data.
Next we square the error function and apply Jensen's inequality to get
\begin{align*}
    \hat{r}_{\mathrm{PL}}(v)^2 =  \Big(\E[\muoh(V, Z) - \muo(V, Z) \mid V = v, \Wfirst]\Big)^2 \leq \E\Big[ \Big(\muoh(V, Z) - \muo(V, Z)\Big)^2 \mid V = v, \Wfirst \Big]
\end{align*}

Taking the expectation over $\Wfirst$ on both sides, we get
\begin{align*}
    \E [\hat{r}_{\mathrm{PL}}(v)^2 \mid V = v] &\leq  \E \Bigg[ \E\Big[\Big(\muoh(V, Z) - \muo(V, Z)\Big)^2 \mid V = v, \Wfirst\Big] \mid V = v \Bigg] \\
    &=  \E \Bigg[ \Big(\muoh(V, Z) - \muo(V, Z)\Big)^2 \mid V = v \Bigg]
\end{align*}

Next, under our stability conditions(\textsection~\ref{condition:stab}), we can apply Theorem 1 of \citet{kennedy2020optimal} (stated in the next section for reference) to get the pointwise bound
\begin{align*}
    \E \Bigg[\Big(\plm - \nu(v)\Big)^2 \Bigg] \lesssim & \E \Bigg[\Big(\tilde \nu(v) - \nu(v)\Big)^2\Bigg] +  \E \Bigg[ \Big(\muoh(V, Z) - \muo(V, Z)\Big)^2 \mid V = v\Bigg]
\end{align*}

 Theorem 1 of Kennedy also implies a bound on the integrated MSE of the \pl approach:
\begin{align*}
    & \E\norm{\plm - \nu(v)}^2 \lesssim  \E \norm{\tilde \nu(v) - \nu(v)}^2  + \int_{\mathcal{V}} \E \Big[ (\muoh(V, Z) - \muo(V, Z))^2 \mid V = v\Big] p(v) dv
\end{align*}
\end{proof}

 \subsubsection{Theorem for Pseudo-Outcome Regression (Kennedy)} 
The proofs of Proposition~\ref{thm:pl} and Theorem~\ref{thm:bc} rely on Theorem 1 of \citet{kennedy2020optimal} which we restate here for reference. 
In what follows we provide the proof for Theorem~\ref{thm:bc}.

\begin{theorem}[Kennedy]\label{thm:kennedy}
Recall that $\Wfirst$ denotes our $n$ first-stage training data samples. 
Let $\hat f(w) := \hat f(w; \Wfirst)$ be an estimate of the function $f(w)$ using the training data $\Wfirst$.
Denote an independent sample as $W$.
The true regression function is $m(v) := \E[f(W) \mid V = v]$.
Denote the second stage regression as ${\hat m(v) := \hat \E_n[\hat f(W) \mid V = v]}$. 
Denote its oracle equivalent (if we had access to $\ya$) as ${\tilde m(v) := \hat \E_n[f(W) \mid V = v]}$.
Under stability conditions(\textsection~\ref{condition:stab}) on the regression estimator $\hat \E_n$, we have the following bound on the pointwise MSE:
\vspace{-0.1pt}
$$ \E \Big[\Big(\hat m(v) - m(v)\Big)^2 \Big]  \lesssim \E \Big[\Big(\tilde m(v) - m(v)\Big)^2\Big] + \E \Big[ \hat r(v)^2 \Big] $$
where $\hat r(v)$ describes the error function $ \hat r(v) := \E[\hat f(W) \mid V = v, \Wfirst] - m(v) $.
This implies the following bound for the integrated MSE:
$$\E\norm{\hat m(v) - m(v)}^2 \lesssim \E \norm{\tilde m(v) -m(v)}^2 + \int \E \big[ \hat r(v)^2 \big]p(v) dv$$
\end{theorem}

\subsubsection{Proof of Theorem~\ref{thm:bc} (error of the \bc method)} \label{sec:proof_bc}
Here we provide the proof for our main theoretical result which bounds the error of our proposed \bc method.
\begin{proof}
As for the \pl error bound above, the first step is to derive the form of the error function for our \bc approach.
For clarity and brevity, we denote the measure of the expectation in the subscript.
\begin{align*}
    \hat r_{\mathrm{DR}}(v) &= \E_{W \mid V= v, \Wfirst} \Bigg[\frac{\indA}{\pih(v,Z)}(Y - \muoh(v, Z)) + \muoh(v, Z)\Bigg] - \nu(v) \\
    &= \E_{Z, A \mid V= v, \Wfirst} \Bigg[ \E_{W \mid A = a, V= v, Z = z, \Wfirst} \Bigg[\frac{\indA}{\pih(v,Z)}(Y - \muoh(v, z)) + \muoh(v, z)\Bigg] \Bigg] - \nu(v) \\
    &= \E_{Z ,A \mid V= v, \Wfirst} \Bigg[ \E_{Y \mid A = a, V= v, Z = z, \Wfirst} \Bigg[\frac{\indA}{\pih(v,Z)}(Y - \muoh(v, z))\Bigg]  + \muoh(v, Z)\Bigg] - \nu(v) \\
    &= \E_{Z, A \mid V= v, \Wfirst} \Bigg[  \frac{\indA}{\pih(v,Z)}(\E_{Y \mid A = a, V= v, Z = z, \Wfirst}[Y] - \muoh(v, Z)) + \muoh(v, Z)\Bigg] - \nu(v) \\
    &= \E_{W \mid V= v, \Wfirst} \Bigg[  \frac{\indA}{\pih(v,Z)}(\muo(v,Z) - \muoh(v, Z)) + \muoh(v, Z)\Bigg] - \nu(v) \\
    &= \E_{Z \mid V= v, ,\Wfirst} \Bigg[ \E_{W \mid V= v, Z = z, \Wfirst} \Bigg[\frac{\indA}{\pih(v,Z)}(\muo(v,z) - \muoh(v, z)) + \muoh(v, z)\Bigg] \Bigg] - \nu(v) \\
    &= \E_{Z \mid V= v, \Wfirst} \Bigg[ \frac{\p(A =a \mid V= v, Z = z)}{\pih(v,Z)}(\muo(v,Z) - \muoh(v, Z)) + \muoh(v, Z) \Bigg] - \nu(v) \\
    &= \E_{Z \mid V= v, \Wfirst} \Bigg[ \frac{\pi(v, Z)}{\pih(v,Z)}(\muo(v,Z) - \muoh(v, Z)) + \muoh(v, Z) \Bigg] - \nu(v) \\
    &= \E_{Z \mid V= v, \Wfirst} \Bigg[ \frac{\pi(v, Z)}{\pih(v,Z)}(\muo(v,Z) - \muoh(v, Z)) + \muoh(v, Z) - \muo(v, Z) \Bigg] \\
    &= \E \Bigg[ \frac{(\muo(v,Z) - \muoh(v,Z))(\pi(v,Z) - \pih(v,Z))}{ \pih(v,Z)} \mid V = v, \Wfirst \Bigg]
\end{align*}

Where the first line holds by definition of the error function $\hat r$ and the second line by iterated expectation.
The third line uses the fact that conditional on $Z = z, V =v, A = a$, then the only randomness in $W$ is $Y$ (and therefore $\muoh$ is constant).
The fourth line makes use of the $(\indA)$ term to allow us to condition on only $A = a$ ( since the term conditioning on any other $a' \neq a$ will evaluate to zero).
The fifth line applies the definition of $\muo$. \\
The sixth line again uses iterated expectation and the seventh makes use of the fact that conditional on $Z$, the only randomness now is in $A$ and that $\Wfirst$ is an independent randomly sampled set.
The seventh line applies the definition of ${\pi(v,z) = \mathbb{P}(A = 1 \mid V= v, Z = z)}$ which since $A \in \{0,1\}$ is equal to $\E[A \mid V = v, Z = z]$. 
The eight line uses iterated expectation and the fact that $\Wfirst$ is an independent randomly sampled set to rewrite ${\nu(v) = E_{Z \mid V =v, \Wfirst } [\muo(v, Z)}] $. 
The ninth line rearranges the terms.

By Cauchy-Schwarz and the positivity assumption,
\begin{align*}
   &\hat{r}_{\mathrm{DR}}(v)\leq C \sqrt{\E[(\muo(v,Z) - \muoh(v,Z))^2 \mid V = v, \Wfirst]}\sqrt{\E[(\pi(v,Z) - \pih(v,Z))^2 \mid V = v, \Wfirst]}
\end{align*}
for a constant $C$. 

Squaring both sides yields
\begin{align*}
   &\hat{r}_{\mathrm{DR}}^2(v)\leq C^2 \ \E[(\muo(v,Z) - \muoh(v,Z))^2 \mid V = v, \Wfirst] \ \E[(\pi(v,Z) - \pih(v,Z))^2 \mid V = v, \Wfirst]
\end{align*}

If $\pih$ and $\muoh$ are estimated using separate training samples, then taking the expectation over the first-stage training sample $\Wfirst$ yields:
\begin{align*}
   &\E[\hat{r}_{\mathrm{DR}}^2(v)] \leq C^2 \ \E[(\muo(v,Z) - \muoh(v,Z))^2] \mid V = v] \ \E[(\pi(v,Z) - \pih(v,Z))^2] \mid V = v]
\end{align*}

Applying Theorem 1 of \citet{kennedy2020optimal} gets the pointwise bound: 
\begin{align*} 
  \E \Bigg[\Big(\bcm - \nu(v)\Big)^2 \Bigg] \lesssim & \  \E \Bigg[\Big(\tilde \nu(v) - \nu(v)\Big)^2\Bigg] \\
  &+ \E \Big[ (\pih(V, Z) - \pi(V, Z))^2 \mid V = v\Big] \E \Big[ (\muoh(V, Z) - \muo(V, Z))^2 \mid V = v\Big]
\end{align*}

and the bound on integrated MSE of the \bc approach:
\begin{align*}
    & \E\norm{\bcm - \nu}^2 \lesssim \  \E \norm{\tilde \nu(v) - \nu(v)}^2 
    \\ &+ \int_{\mathcal{V}} \E \Big[ (\pih(V, Z) - \pi(V, Z))^2  \mid V = v\Big] \E \Big[ (\muoh(V, Z) - \muo(V, Z))^2 \mid V = v\Big] p(v)dv
\end{align*}

\end{proof}

\subsection{Efficient influence function for \bc method}
We provide the efficient influence function of the \bc method.
The efficient influence function indicates the form of the bias-correction term in the \bc method.
The efficient influence function $\phi(A, V, Z, Y)$
for parameter $ {\psi(V) :=  \E[ \ya \mid V] = \E[ \E[ Y \mid V, Z, A =a] \mid V]}$ is
\begin{equation*}
   \phi(A, V, Z, Y) = \frac{\mathbb{I}\{A = a\}}{\pi(V,Z)}(Y - \muo(V,Z)) + \muo(V,Z) - \psi(V) 
\end{equation*}


\section{Synthetic experiment details and additional results}

In this section we present details on the synthetic experiments and present additional results.
We present the random forests graphs omitted from the main paper,  results on calibration-type curves that show where the errors are distributed, and experiments on our evaluation procedure.


\subsection{Experimental details}

\paragraph{More details on data-generating process}
We designed our data-generating process in order to simulate a real-world risk assessment setting. 
We consider both $V$ and $Z$ to be risk factors whose larger values indicate increased risk and therefore we construct $\mu$ to increase with $V$ and $Z$.
Our goal is to assess risk under the null (or baseline) treatment as per \cite{coston2020counterfactual}, and we construct $\pi$ such that historically the riskier treatments were more likely to get the risk-mitigating treatment and the less risky cases were more likely to get the baseline treatment.

We now provide further details on the choices of coefficients and variance parameters.
In the first set of experiments presented in the main paper, we simulate $V_i$ from a standard normal, and in the uncorrelated setting (where $\rho = 0$) we also simulate $Z_i$ from a standard normal. 
In the correlated setting, we sample $Z_i$ from a normal with mean $\rho V_i$ and variance $1- \rho^2$ so that the Pearson's correlation coefficient between $V_i$ and $Z_i$ is $\rho$ and so that the variance in $Z_i = 1$. 
We simulate $\mu$ to be a sparse linear model in $V$ and $Z$ with coefficients of 1 when $\rho = 0$.
When $\rho \neq 1$, the coefficients are set to $\frac{\kmuv}{\kmuv+\rho \kmuz}$ so that the $L_1$ norm of the $\nu$ coefficients equals $\kmuv$ for all values of $\rho$. 
Without this adjustment, changing $\rho$ would impact error by also changing the signal-to-noise ratio in $\nu$.
We simulate the potential outcome $\ya$ to be conditionally Gaussian and the choice of variance $\frac{1}{2n}\norm{\muo(V,Z)}_2^2 $ yields a signal-to-noise ratio of 2.
The specification for $\nu$ follows from the marginalization of $\mu$ over $Z$.
The propensity score $\pi$ depends on the sigmoid of a sparse linear function in $V$ and $Z$ that uses coefficients $\frac{1}{\sqrt{\kmuv + \kmuz}}$ in order to satisfy our positivity condition. 

We use $d = 500$, $n = 1000$, $\kmuv = 25$, and $0 \leq \kmuz \leq 45$ to simulate a sparse high-dimensional setting with many measured variables in the training data, of which only 5\%-15\% are predictive of the outcomes. 
In one set of experiments, we vary the value of $\kmuz$ to assess impact of various levels of confounding on performance.
In other experiments, where we vary $\rho$ or the dimensionality of $V$ ($\dv$), we use $\kmuz = 20$ so that $V$ has slightly more predictive power than the hidden confounders $Z$.

\paragraph{Hyperparameters} Our LASSO presents are presented for cross-validated hyperparameter selection using the \texttt{glmnet} package in R. 
The random forests results use $1000$ trees and default \emph{mtry} and splitting parameters in the \texttt{ranger} package in R.

\paragraph{Training runs}
Defining a training run as performing a learning procedure such as LASSO, for a given hyperparameter selection and given simulation, the \btr method trains in one run, the \pl method trains in two runs, and the \bc method trains in three runs. 
For a given simulation, the exact number of runs depends on the hyperparameter tuning.
Since we only ran random forests (RF) for the default parameters,  the \btr method with RF trained in one run, the \pl method with RF trained in two runs, and the \bc method with RF trained in three runs. 
The LASSO results using \texttt{cv.glmnet} were tuned over $\leq 100$ values of $\lambda$; the \btr method with LASSO trained in $\leq 100$ runs, the \pl method with LASSO trained in $\leq 200$ runs, and the \bc method with LASSO trained in $\leq 300$ runs.

\paragraph{Sample size and error metrics}
For experiments in the main paper, we trained on $n= 1000$ datapoints. 
We test on a separate set of $n = 1000$ datapoints and report the estimated mean squared error (MSE) on this test set using the following formula:

$$ \frac{1}{n} \sum_{i=1}^n (\nu(V_i) - \hat \nu(V_i))^2 $$ 


\paragraph{Computing infrastructure}

We ran experiments on an Amazon Web Serivces (AWS) c5.12xlarge machine.
This parallel computing environment was useful because we ran thousands of simulations. 
The traintime of each simulation, entailing the LASSO and RF experiments, took 1.8 seconds. 
In practice for most real-world decision support settings, our method can be used in standard computing environments; relative to existing predictive modeling techniques, our method will require $\leq 3X$ the current train time.
Our runtime depends only on the regression technique used in the second stage and should be competitive to existing models.

\subsection{Random forest results}
\begin{figure}
\centering
      \includegraphics[width=0.25\linewidth]{img/legend.pdf}\\
      \includegraphics[width=0.03\textwidth, trim=300 0 220 0]{img/axis.pdf}
    \begin{subfigure}[t]{0.40\textwidth}
        \centering
         \includegraphics[width=\linewidth, trim=23 20 14 10]{img/uncor/vary_zeta_p400_RF.pdf}
        \caption{Random forests}
    \end{subfigure}%
    ~
    \begin{subfigure}[t]{0.40\textwidth}
        \centering
        \includegraphics[width=\linewidth, trim=23 20 14 10]{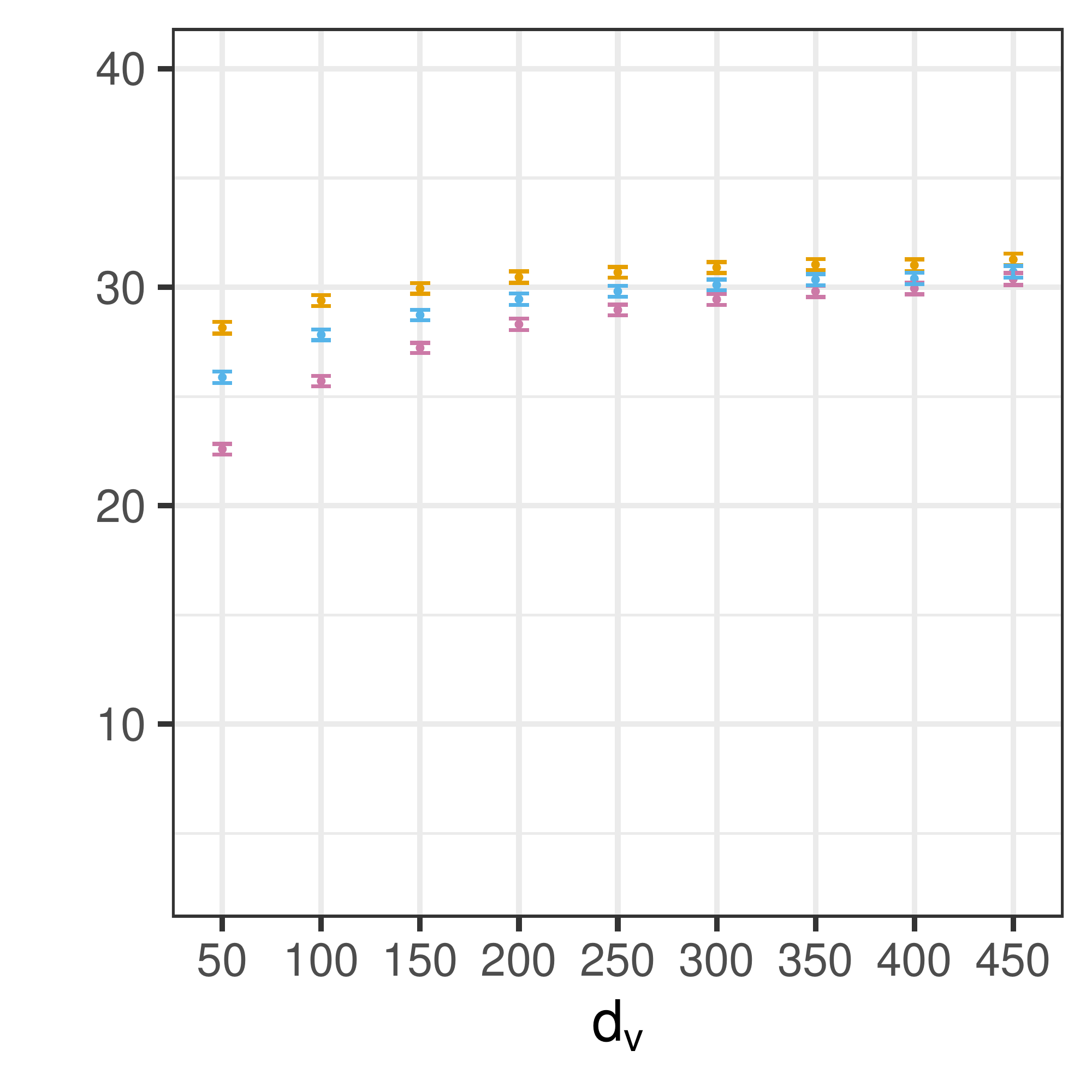}
         \caption{Random forests}
    \end{subfigure}
    \caption{\textbf{(a)} MSE as we vary $\kmuz$ using random forests to learn $\pih$, $\muoh$, $\hat{\nu}_{\mathrm{TCR}}$, $\hat{\nu}_{\mathrm{PL}}$, $\hat{\nu}_{\mathrm{DR}}$ for $\vzcor = 0$,  $\dv = 400$ and $\kmuv = 25$.  \\
    \textbf{(b)} MSE against $\dv$ using random forests and $\vzcor = 0$,  $\kmuv = 25$ and $\kmuz = 20$. \\
     Error bars denote $95\%$ confidence intervals.}
    \label{fig:uncor-rf}
\end{figure}

Figure~\ref{fig:uncor-rf} presents the results when using random forests for the first and second stage estimation in the uncorrelated V-Z setting. 
Figure~\ref{fig:uncor-rf}a was provided in the main paper, and we include it here again for ease of reference.
Figure~\ref{fig:uncor-rf}b shows how method performance varies with $\dv$.
At low $\dv$, the \btr method does significantly better than the two counterfactually valid approaches.
This suggests that the estimation error incurred by the \pl and \bc methods outweighs the confounding bias of the \btr method.

\begin{figure}
\centering
      \includegraphics[width=0.2\linewidth]{img/legend.pdf}\\
      \includegraphics[width=0.03\textwidth, trim=300 0 220 0]{img/axis.pdf}
    \begin{subfigure}[t]{0.31\textwidth}
        \centering
        \includegraphics[width=\linewidth, trim=23 20 14 15]{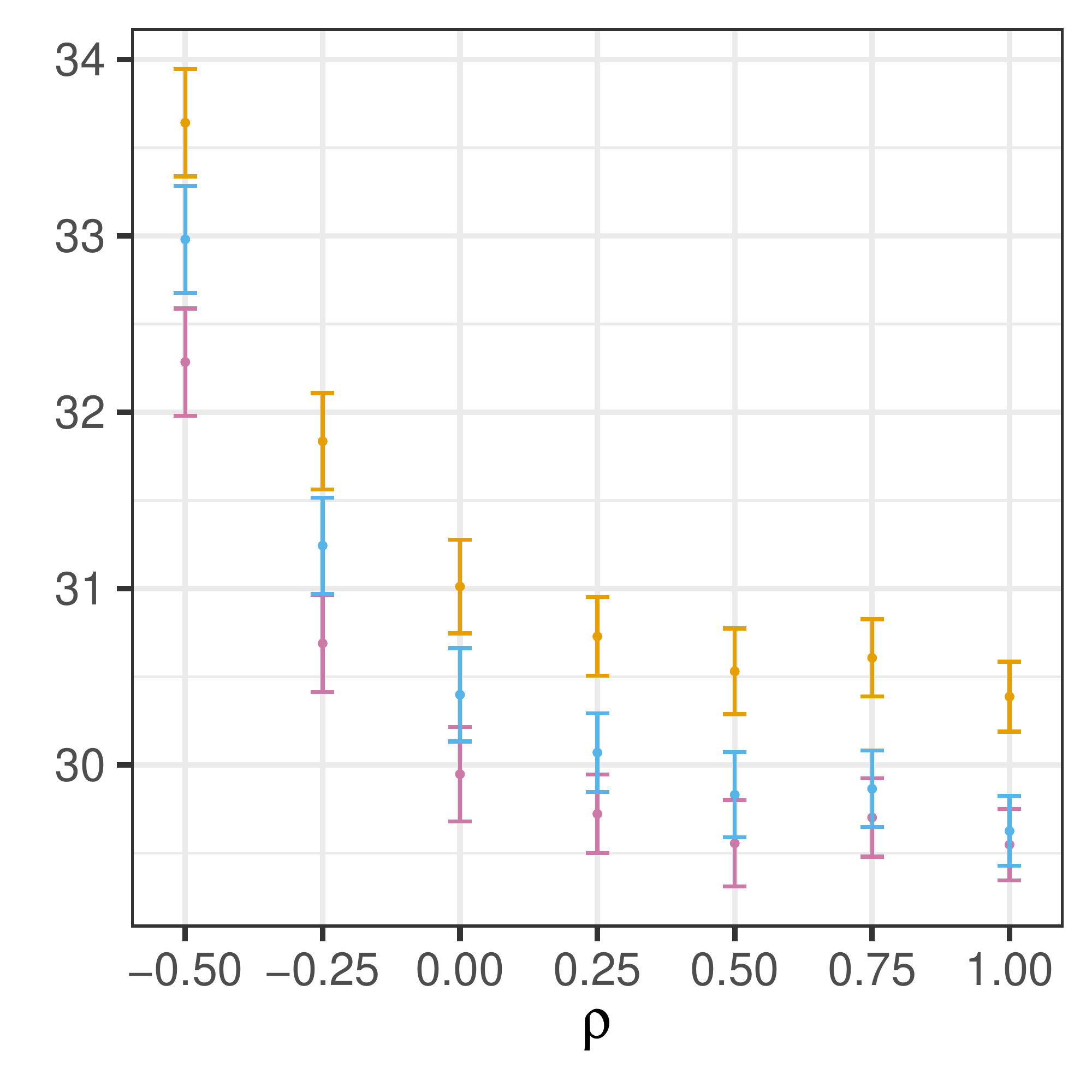}
        \caption{Random forests}
        \end{subfigure}%
    ~
    \begin{subfigure}[t]{0.31\textwidth}
        \centering
         \includegraphics[width=\linewidth, trim=23 20 14 15]{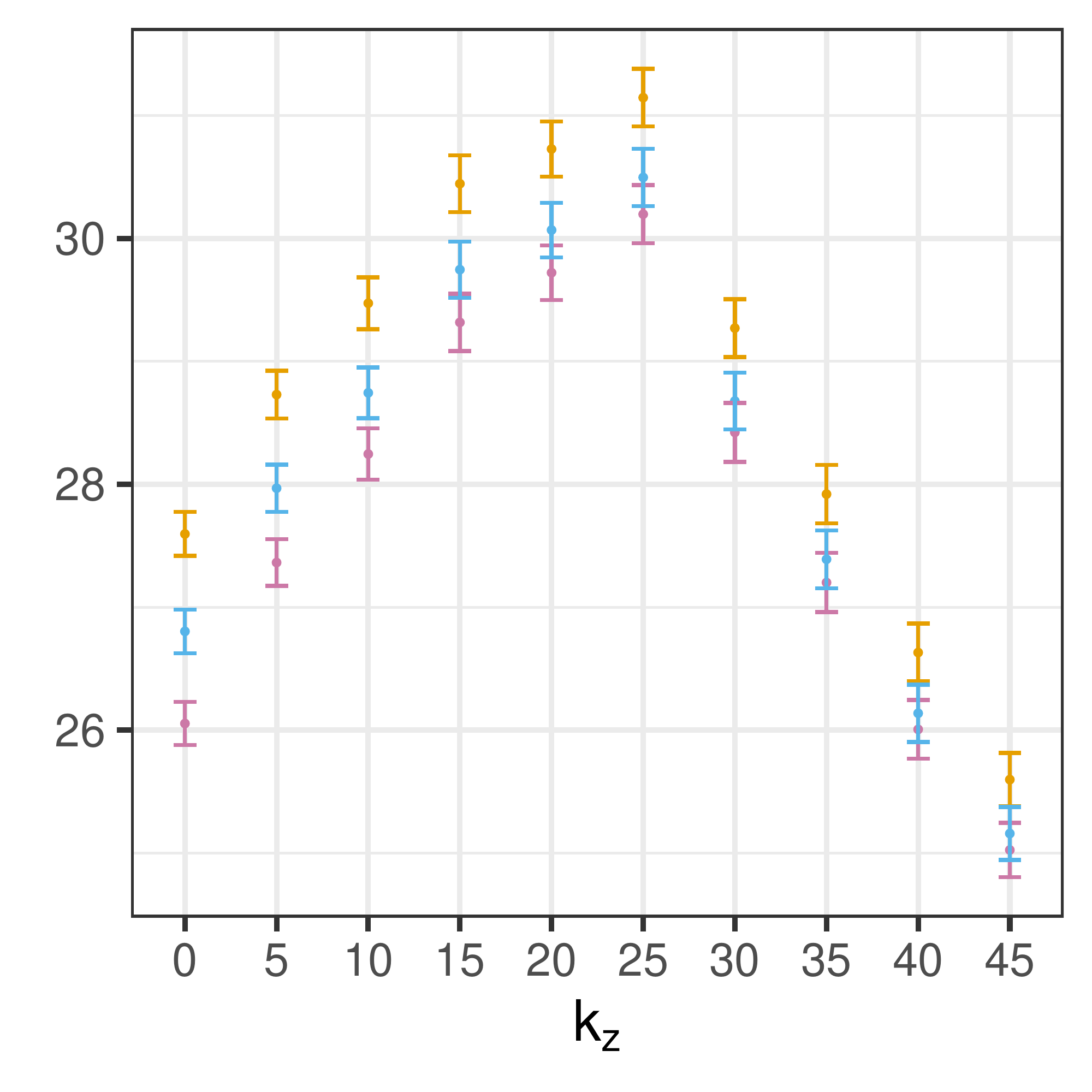}
        \caption{Random forests}
    \end{subfigure}%
    ~
    \begin{subfigure}[t]{0.31\textwidth}
        \centering
        \includegraphics[width=\linewidth, trim=23 20 14 15]{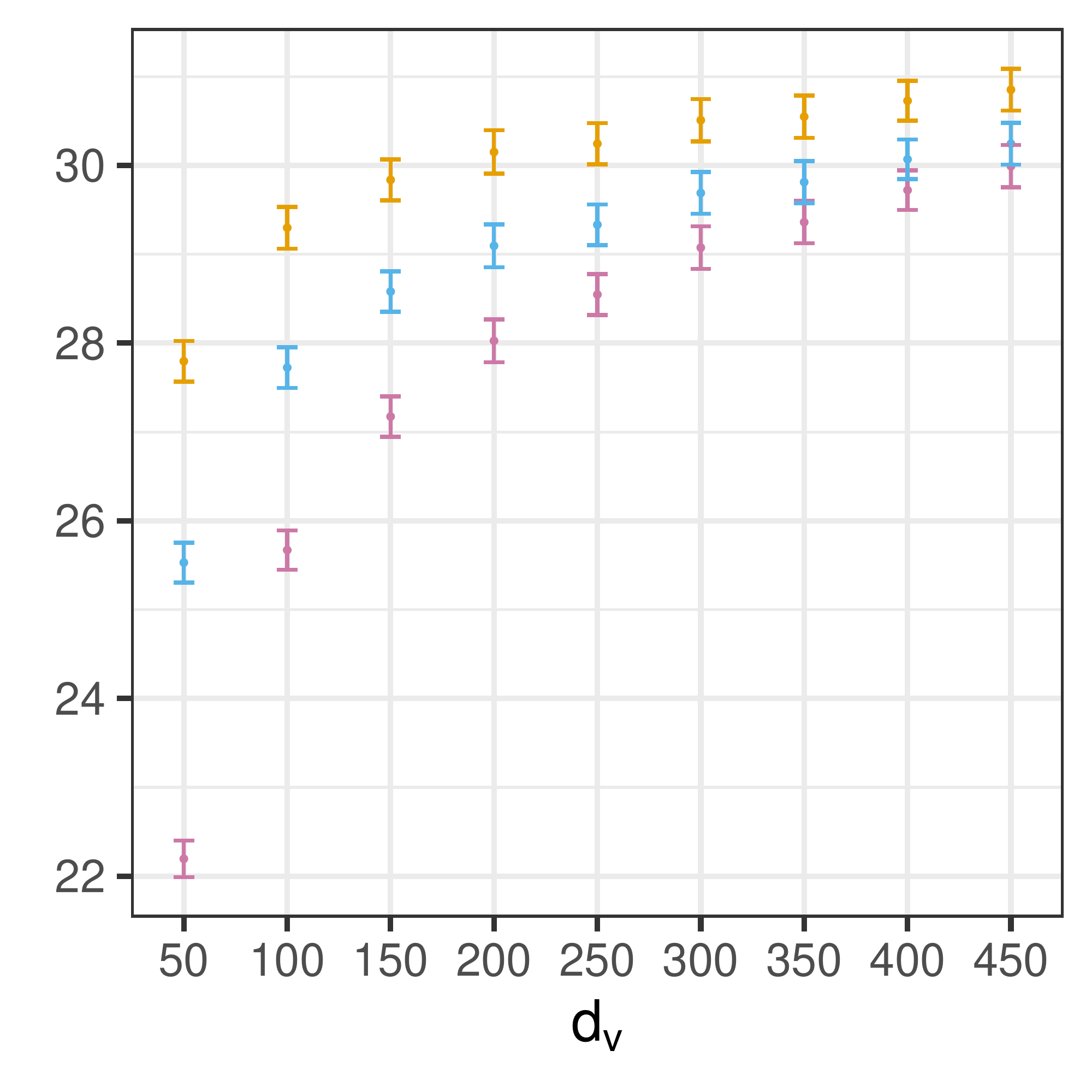}
         \caption{Random forests}
    \end{subfigure}
    \caption{\textbf{(a)} MSE against correlation $\vzcor_{V_i, Z_i}$ for $\kmuz = 20$, $\kmuv = 25$, and $\dv = 400$.
For all methods, error decreases with $\vzcor \leq 0.5$, at which point the error does not change with increasing $\vzcor$.
    \textbf{ (b)} MSE as we increase $\kmuz$ for $\vzcor = 0.25$, $\kmuv = 25$, and $\dv = 400$. Compare to  Figure~\ref{fig:uncor-rf}a; the weak positive correlation reduces MSE, particularly for  $\kmuv < i \leq \kmuz$ when $V_i$ is only a correlate for the confounder $Z_i$ but not a confounder itself. 
    \textbf{(c)} MSE against $\dv$ for $\vzcor = 0.25$, $\kmuz = 20$, and $\kmuv = 25$. 
As with the uncorrelated setting (\ref{fig:uncor-rf}b), the \bc{} and \btr methods are better able to take advantage of low $\dv$ than the \pl{} method. \\
 Error bars denote $95\%$ confidence intervals.
}
    \label{fig:cor-rf}
\end{figure}

\subsection{Evaluation experiments}
To empirically assess our proposed doubly-robust evaluation procedure, we generated one sample of training data with $n= 1000$, $d = 500$, $\dv = 200$, $\kmuv =25$, and $\kmuz = 30$ as well as a "ground-truth" test set with $n = 10,000$. 
We trained the \btr, \pl, and \bc methods on the training data and estimated their true performance on the large test set.
The true prediction error $$\frac{1}{n} \displaystyle \sum_{i=1}^n \big(Y_i^a - \hat{\nu}(V_i)\big)^2$$ was 77.53, 74.12, and 72.68 respectively for the \btr, \pl and \bc methods.
We then ran 100 simulations where we sampled a more realistically sized test set of $n = 2000$.
In each simulation we performance the evaluation procedure to estimate prediction error on the observed data. 
The MSE estimator  with $95\%$ CI covered the true MSE 94 times for the DR approach and 93 times for the PL. 
81\% of the simulations correctly identified the \bc procedure as having the lowest error, $14\%$ suggested that the \pl procedure had the lowest error and $5\%$ suggested that the \btr had the lowest error.

For additional experimental results on using doubly-robust evaluation methods for predictive models, we recommend \cite{coston2020counterfactual}.

\subsection{Calibration-styled analysis of the error}
\begin{figure}
\centering
      \includegraphics[width=0.15\linewidth]{img/legend.pdf}\\
\begin{subfigure}[t]{0.5\textwidth}
    \centering
        \includegraphics[width=\linewidth, trim=23 20 14 2]{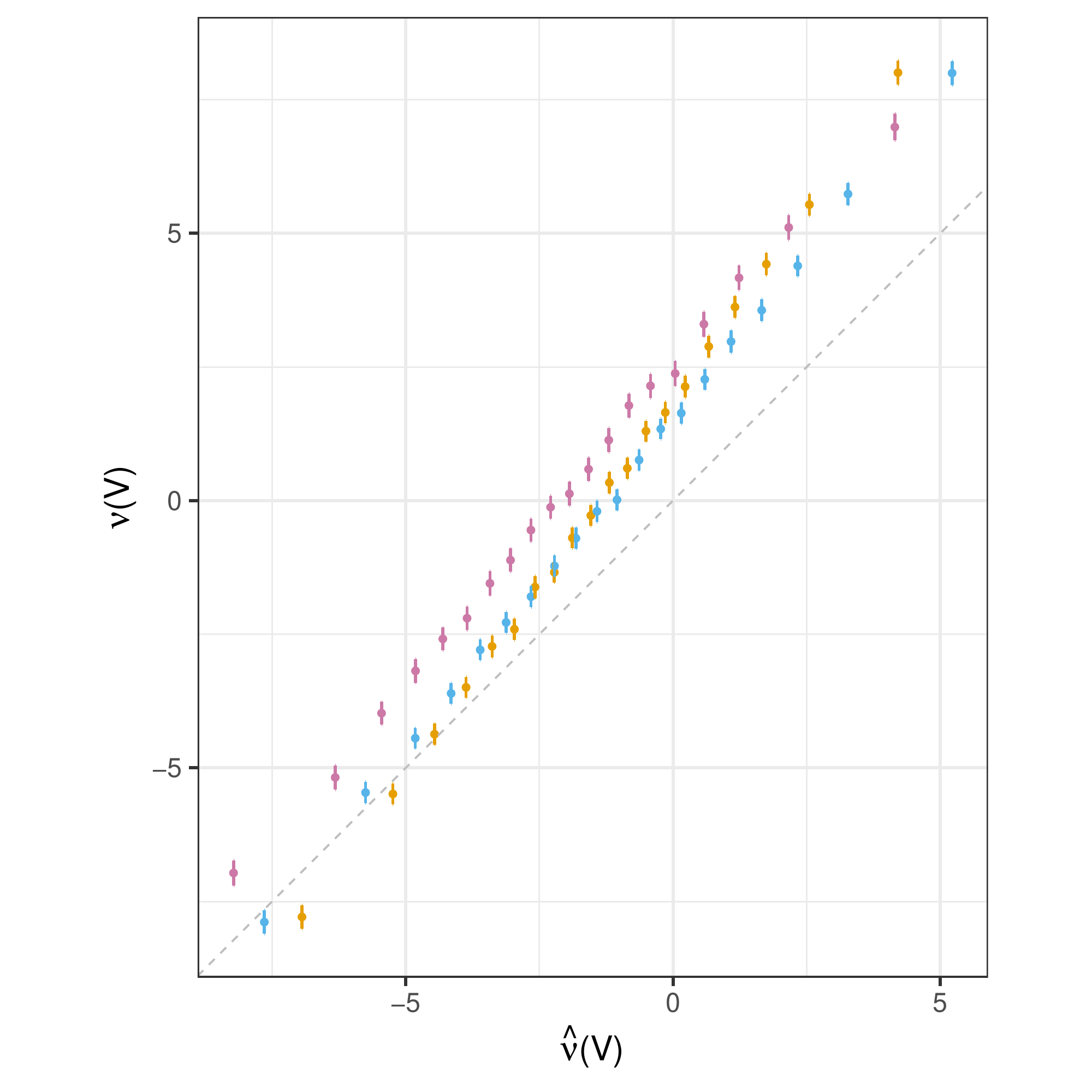}
    \caption{LASSO}
    \end{subfigure}%
    ~
    \begin{subfigure}[t]{0.5\textwidth}
        \centering
        \includegraphics[width=\linewidth, trim=23 20 14 2]{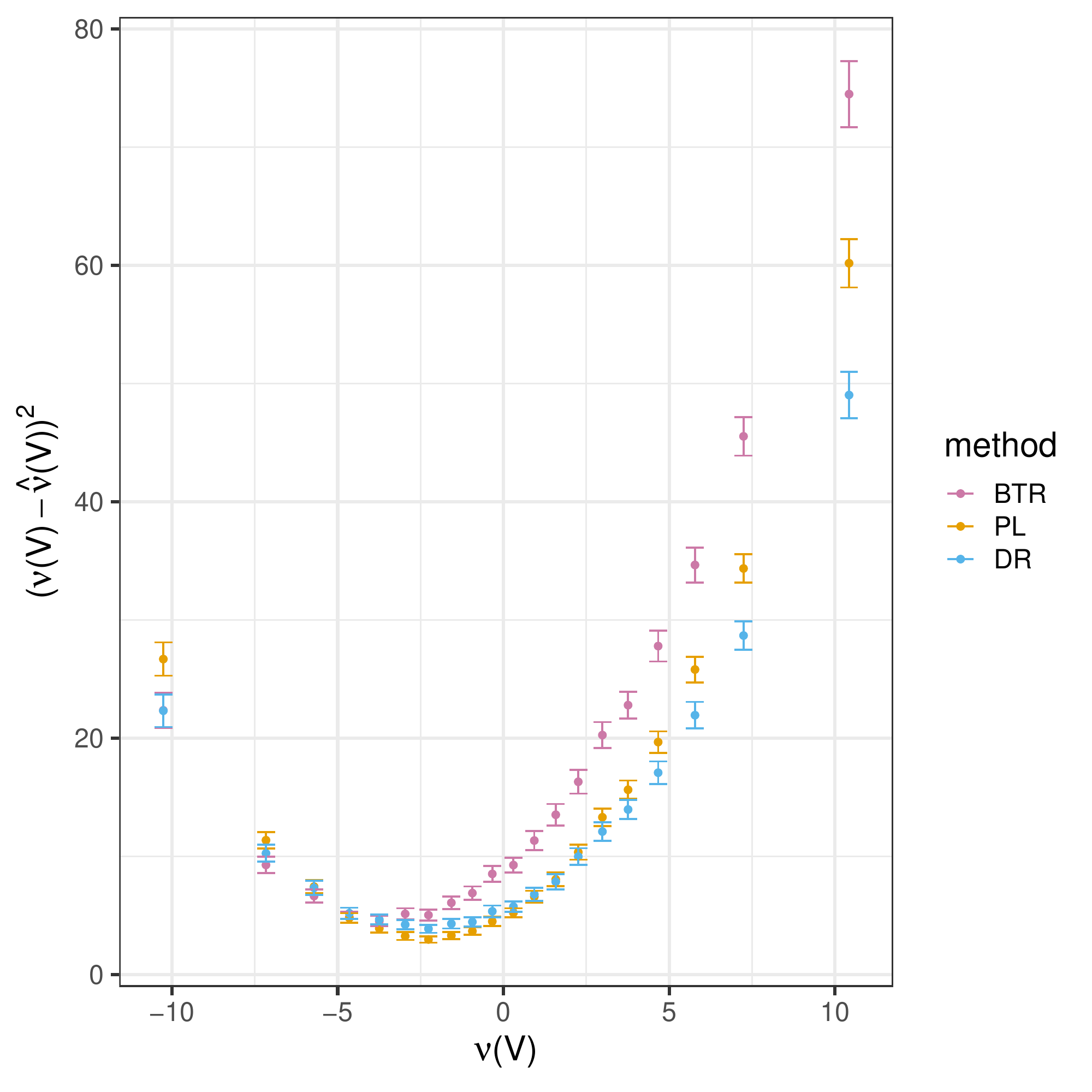}
        \caption{LASSO}
        
    \end{subfigure}
    \caption{(a) Calibration plot for LASSO regressions with $p = 400$, $q = 100$, $\kmuz = 20$ and  $\kmuv = 25$. A well-calibrated model will track the dotted $y = x$ line. Our \bc model is the best calibrated. As expected from its confounding bias, the \btr method underestimates risk for all predicted values. Interestingly the \pl and \bc methods also underestimate risk for higher predicted risk values.   \\ (b) Squared error against true risk $\nu(V)$ for LASSO regressions with $p = 400$, $q = 100$, $\kmuz = 20$ and  $\kmuv = 25$. All models have highest error on the riskiest cases (those with large values of $\nu(V)$); this is particularly pronounced for the \btr model, suggesting that the \btr model would make misleading predictions for the highest risk cases.}
    \label{fig:reg_calib}
\end{figure}

Above  we analytically showed that in a standard risk assessment setting the \btr method underestimates risk. 
We empirically demonstrate this in Figure~\ref{fig:reg_calib} where the calibration curve (Figure~\ref{fig:reg_calib}a) shows that \btr underestimates risk for all predicted values.
Figure~\ref{fig:reg_calib}b plots the squared error against true risk $\nu(V)$, illustrating that errors are extremely large for high-risk individuals, particularly for the \btr model.
This highlights a danger in using confounded approaches like the \btr model: they make misleading predictions about the highest risk cases.
In high-stakes settings like child welfare screening, this may result in dangerously deciding to \emph{not} investigate the cases where the child is at high risk of adverse outcomes \cite{coston2020counterfactual}.
The counterfactually valid \pl and \bc models mitigate this to some effect, but future work should investigate why the errors are still large on high-risk cases and propose procedures to further mitigate this.

\section{Real-world experiment details and additional results}
In this section we elaborate on the details of our evaluation of the methods on a real-world child welfare screening task. 
\subsection{Child welfare dataset details}
We use a dataset of over 30,000 calls to the child welfare hotline in Allegheny County, Pennsylvania.
Each call contains more than 1000 features, including information on the allegations in the call as well as county records for all individuals associated with the call.
The call features are categorical variables describing the allegation types and worker-assessed risk and danger ratings. 
The county records include demographic information such as age, race and gender as well as criminal justice, child welfare, and behavioral health history. 
The outcome we wish to predict is whether the family would be offered services if the case were screened in for investigation. 

\subsection{Child welfare experimental details}

We perform the first stage regressions using random forests to allow us to flexibly estimate the nuisance function $\pi$ and $\mu$. For the second stage regressions, we use LASSO to yield interpretable prediction models.

\paragraph{Hyperparameters} 
The first stage random forest regressions use $500$ trees and the default \emph{mtry} and splitting parameters in the \texttt{ranger} package in R.
For our LASSO second stage regressions, we use cross-validation in the \texttt{glmnet} package in R to select the LASSO penalty parameters.

\paragraph{Training runs}
Each of the two nuisance function estimations in the first stage trains in one run.
The LASSO cross-validation using \texttt{cv.glmnet} tunes over $\leq 100$ values of $\lambda$.
 Therefore,
the \btr method  trains in $\leq 100$ runs, the \pl method with LASSO trains in $\leq 101$ runs, and the \bc method with LASSO trains in $\leq 102$ runs.

\paragraph{Sample size and error metrics}
The dataset consists of 30,000 calls involving over 70,000 unique children. We partitioned the children into train and test partitions using a graph partitioning procedure that ensured that all siblings were contained within the same partition to avoid the contamination problem discussed in \cite{chouldechova2018case}.
In order to enable more precise estimation of the counterfactual outcomes in this real-world setting, we perform a 1:2 train-test split such that the train split contains 27000 unique children and the test split contains 50000 unique children. 
We use the evaluation procedure in \textsection~\ref{sec:eval} to obtain estimates of the MSE with confidence intervals.

\paragraph{Computing infrastructure}
All real-world experiments were run on a MacBook Pro with an 8-core i9 processor and 16 GB of memory.
Each first stage regression trained in 15 seconds. Each second stage regression trained in 4.5 minutes.

\subsection{Modeling human decisions}
Algorithmic tools used in decision support settings often estimate the likelihood of an event (outcome) under a proposed decision. 
This is the setting for which our method is tailored.
By contrast, another paradigm trains algorithms to predict the human decision. 
We present here the results of such an algorithm when evaluated against the downstream outcome of interest (services offered).
To train this model, we used the historical screening decision as the outcome. 
We allowed this model to access all confounders (both $V$ and $Z$, as if we did not have runtime confounding), yet this approach achieves a significantly higher MSE of $0.3207$ with $95\%$ confidence interval $(0.3143, 0.3271)$. 
It should not be surprising that a model trained on human decisions performs worse than models trained on downstream outcomes when we are evaluating against the downstream outcomes. 
This highlights the importance of using downstream outcomes in decision support settings when the goal is related to the downstream outcome e.g. to mitigate the risk of a downstream outcome or to prioritize cases that will benefit from the decision treatment.